\documentclass[fleqn,12pt]{SelfArx} 

\usepackage[english]{babel} 

\definecolor{color1}{RGB}{0,0,90} 
\definecolor{color2}{RGB}{0,20,20} 

\usepackage{hyperref} 

\hypersetup{
	hidelinks,
	colorlinks,
	breaklinks=true,
	urlcolor=color2,
	citecolor=color1,
	linkcolor=color1,
	bookmarksopen=false,
	pdftitle={Title},
	pdfauthor={Author},
}

\usepackage{wrapfig}
\usepackage{gensymb}
\usepackage{amsmath}

\usepackage{setspace}

\JournalInfo{May 2024} 
\Archive{} 

\PaperTitle{Improvement and Empirical Testing of a Novel Autonomous Microplastics-Collecting Semisubmersible} 

\Authors{Ziddane Isahaku\textsuperscript{1}*} 
\affiliation{\textsuperscript{1}\textit{Nicolet High School, Milwaukee, Wisconsin, United States}} 
\affiliation{*\textbf{Corresponding author}: ziddaneisahaku@gmail.com} 

\newcommand{\keywordname}{Keywords} 

\Abstract{Since their invention, plastics have become ubiquitous in modern societies all around the world, and their impact on the environment has, in recent years, become nearly as well-known. Plastics produced by humans have reached nearly every corner of the world, and throughout their centuries-long lifetimes, plastics continually break down into smaller and smaller particles due to the physical stresses which they are subjected to. These stresses eventually, inevitably, break these plastics down into microplastics –pieces of plastic small enough to be consumed by organisms in bodies of water throughout the globe. These microplastics can very easily bioaccumulate, and have been found everywhere from the Great Lakes to the bloodstreams of humans. The effects of these plastics are poorly understood, however, they have been linked to infertility, halted growth, and a host of other maladies in aquatic organisms. Currently, removal of these plastics has been neglected, with no governmental action to remove them from marine environments, and this project aims to begin prototyping a solution to this issue. A significant percentage of microplastics are found at the surface of waterways, thus trawling in surface waters using an autonomously propelled net is proposed as a way to solve this seemingly intractable issue. By attaching motors and a guidance system to a manta trawl, a device currently used for collecting microorganisms, the process of collecting microplastics in open water can be automated, and thus the work of removing plastics from the environment on a large scale can begin.}

\begin{document}

\maketitle 

\tableofcontents 

\listoffigures

\thispagestyle{empty} 


\section{Acknowledgments}
\begin{itemize}
    \item I would like to thank Mrs. Stephanie Rasmussen of Nicolet High School for her guidance and advice throughout the course of this project.
    \item I would like to thank Mr. Adam Thiel and Nicolet High School for the use of school facilities when manufacturing some parts of this project.
    \item I would like to thank Mr. Peter L. Lenaker of USGS for his advice regarding the environmental effects of microplastics as well as for the provision of a Manta Trawl for modification in this project.
    \item I would like to thank Dr. Wei Wang of MIT for his advice in manufacturing and design of submersibles and power systems.
    \item I would like to thank Mr. Matt Leuker of the University of Minnesota for advice on how to best conduct computational fluid dynamics analyses of the trawl.
    \item Additionally, I would like to thank: Noah Jankowski, Golan Altman-Shafer, and
    \item Jacob Byers for help in collecting and counting microplastics.
    \item Finally, I would like to greatly thank Dr. Gordon G. Parker, Ms. Tania Demonte Gonzalez, and Mr. Vasu Bhardwaj for graciously allowing me to test my device in MTU Wave.
\end{itemize}

\section*{Introduction} 

\addcontentsline{toc}{section}{Introduction} 

\subsection{Rationale}
    The purpose of this project is to create a submersible capable of lowering the quantity of microplastics in waterways via filtration in order to reduce the effects of biomagnification of microplastics in animals and in order to reduce the quantity of microplastics within natural products consumed by humans. Biomagnification, in this case, is the process by which small amounts of microplastics in waterways are consumed by small animals, which are then consumed by humans, leading to accumulation of the pollutant in the body of humans and aquatic animals. Thus there are three primary reasons why this solution has been proposed:
 \begin{enumerate}
     \item Microplastic pollution is an issue which has been observed (especially within the Great Lakes) to be a widespread and harmful phenomenon which negatively affects both the health of aquatic organisms and the health of humans consuming them.
     \item Microplastics are readily uptaken by many of the foods like fish which humans eat regularly, and once microplastics enter an organism, they have been shown to slow or halt growth, cause cellular inflammation, and promote conditions such as obesity.
     \item Filtration through the use of a manta trawl has been proven in past studies to collect microplastics from aquatic environments consistently and effectively.
 \end{enumerate}
	Thus reducing the quantity of microplastics which are taken up by small organisms who are primary and secondary consumers in aquatic ecosystems (through filtration) can significantly reduce biomagnification of microplastics, and can limit their effects on humans as well as other organisms. This has been suggested to be viable by previous studies, which have been able to collect plastic from the Great Lakes and other areas at a significant rate. Further promoting mobile filtration as a way to remove surficial microplastics, studies have suggested that roughly ½ of microplastics fall to the benthic (seafloor) regions of lakes and rivers. Thus collecting a much smaller amount of plastic than exists in totality could significantly lower concentration of plastics in water, which would significantly affect the amount of plastic actually being consumed and bioaccumulated. As a note, despite this potentially easing the workload of a mobile filtration system such as that proposed, microplastics do not cease to be an issue once at the bottom of a body of water, and collecting negatively buoyant microplastics is a significant issue in and of itself. 

\subsection{Definition of Terms Used}

\begin{enumerate}
    \item Ultrasonic Sensor
    \subitem A sensor which utilizes sound waves outside of the spectrum of human hearing in order to measure distance.
    \item CAD
    \subitem Computer Aided Design, a process by which to plan and create components digitally.
    \item Arduino
    \subitem Microcontroller that can be programmed in C to perform various tasks.
    \item Microcontroller
    \subitem A device capable of executing tasks through the use of electrical and software engineering.
    \item C
    \subitem A low-level programming language used (in this instance) for interfacing with an Arduino microcontroller.
    \item Array
    \subitem An array is a data structure consisting of a collection of elements (values or variables), each identified by at least one array index.
    \item Moving Average
    \subitem A succession of averages derived from successive segments (typically of constant size and overlapping) of a series of values.
    \item IDE
    \subitem Integrated development environment: a single application or site allowing for program development.
    \item PWM
    \subitem Pulse Width Modulation, a method of sending complicated messages with a binary output. This is done through the use of on/off signals spaced apart by very short pauses (typically in the range of 1000-2000 microseconds).
    \item ESC
    \subitem Electronic Speed Controller, a device used to control the speed of a brushless motor based on a PWM input.
    \item Microplastics
    \subitem Small pieces of plastic with a diameter under 5mm.
    \item USGS
    \subitem Acronym for the United States Geological Survey, an agency focused on documenting and measuring various natural resources around the US.
\end{enumerate}

\section{Review of Past Literature}
\subsection{Microplastic Effects \& Distribution}
\subsubsection{Effect on Humans}
\paragraph{Ranking of potential hazards from microplastics polymers in the marine environment}
    The purpose of Ranking of potential hazards from microplastics polymers in the marine environment is primarily to evaluate the risks posed by the different types of plastic commonly found in marine environments throughout the world. The authors analyze this danger based upon a variety of factors, including but not limited to their quantity, distribution, biodegradability, density, and monomer toxicity (the toxicity of unpolymerized monomers). The authors found that the plastics PUR(Polyurethane), PVC(Polyvinyl Chloride), PAN(Polyacrylonitrile), ABS(Acrylonitrile Butadiene Styrene), and PMMA (Polymethyl Methacrylate) were the top 5 most dangerous plastics (in descending order) because of their statistics in the aforementioned categories. They also found that PC(Polycarbonate), PF(Phenol Formaldehyde resin), and PP(Polypropylene) were among the least dangerous plastics \cite{Yuanhazards}. 
    This paper was used in the rationale to justify the project as well as to quantify the risk posed with the microplastics which the project focuses on. In addition to this, the relative dangers of the different 3d-printing plastics ABS and PLA were considered in the construction of the device, as both had waterproof properties which are useful for this project. This paper’s information was also used when discussing the project with professionals as a reference for background information on the dangers of microplastics in conjunction with the study Surface Pattern Analysis of Microplastics and Impact on Human Derived Cells.

\paragraph{Surface pattern analysis of microplastics and impact on human derived cells}
    This study focuses on measuring the effects of ABS and PVC microplastics on human derived cells based on their shapes and composition. These shapes included spherical, filament, fragment, and others, and the microplastics were produced by placing the plastics inside of a rotating ball mill. It was found that more spherical microplastics are more easily taken up into cells and circulated, while it was easier for filaments to remain in cells and bioaccumulate once taken up. While in or near cells, these plastics can occasionally break the cell walls of the cells surrounding them, however they more often trigger an immune response due to shedding plasticizers like phthalates, esters, and benzoates within them which produced a roughly 20\% decrease in cell viability over a 5-day period in certain types of cells. These plasticizers have also been linked to many of the harmful side effects of microplastics such as infertility, obesity, and diabetes. \cite{Hansurfacepatternanal}. 
    In this project, the information from the paper was used primarily to supplement the rationale of the paper, and to justify the creation of the device by quantifying the harm affected by the microplastics which are currently found in the waters of lakes like Michigan. This paper also gives a better understanding of the significance of the issue, with specific measurements on the effects of different types of plastics like PVC and ABS which are commonly found in bodies of water. In addition to studying the effects of these different kinds of plastics, the paper also studied which shape classifications of microplastics are more likely to bioaccumulate and cause damage to cells, and it found that filaments (which are more easily captured by nets) bioaccumulate the most and cause the most damage, pointing further to a net being an effective way to collect the most harmful of microplastics.

\subsubsection{Distribution of Microplastics}
\paragraph{Vertical Distribution of Microplastics in the Water Column and Surficial Sediment from the Milwaukee River Basin to Lake Michigan}
    This study focussed on the vertical distribution of microplastics throughout the water column of multiple rivers leading into Milwaukee Harbor and Lake Michigan. The levels of microplastics in each section of the water column were measured using Manta Trawls pulled at several depths below the water’s surface along with 1 sediment sample at each location, making for a total of 96 samples. The samples were separated according to the depth, location, and time that they were collected. The data collected is summarized in the table to the right. The data in the MEP graph was collected in the Milwaukee river in Milwaukee proper, MWW on the Menominee River, KKF in the Kinnickinnic River, INH at the innermost point of Milwaukee harbor, OUH at the outermost point of Milwaukee harbor, and LAK in Lake Michigan \cite{LenakerEtAlvertdist}. 

\begin{wrapfigure}{o}{5cm}
    \includegraphics[width=6cm]{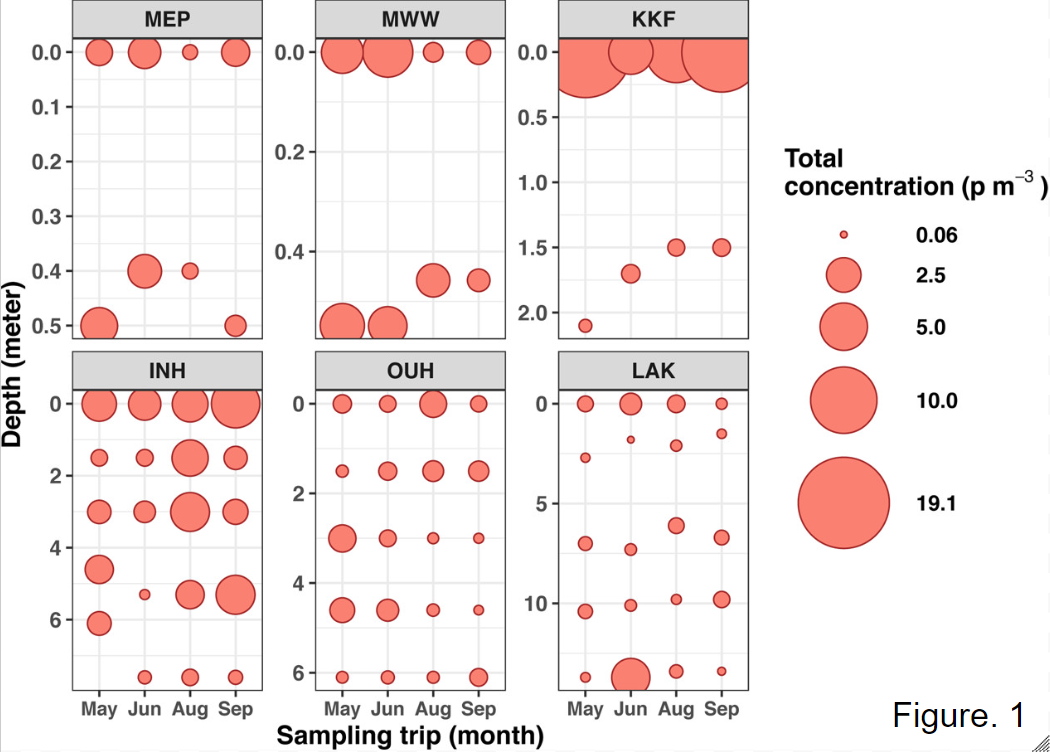}
    \caption{Vertical distribution of microplastics}
\end{wrapfigure}

    The data collected in this study was used primarily to decide upon a depth at which to operate the device created, as well as to provide a reference point for samples collected in any potential testing of the device in order to determine the device’s effectiveness in collecting microplastics. As seen in the graphs, a majority or plurality of the non-benthic microplastics in 19 out of 24 of the locations and times sampled were located between 0 and 0.5 meters of the surface, the range (roughly) collected by manta trawls operating at the surface.

\paragraph{Spatial Distribution of Microplastics in Surficial Benthic Sediment of Lake Michigan and Lake Erie}
    This study focussed on the distribution of microplastics in the benthic sediment of the Great Lakes Erie and Michigan, with 20 samples taken from Lake Michigan and 12 taken from Lake Erie. The measurements collected indicated that the vast majority of benthic (sedimentary) microplastics were fibers and lines, which is consistent with the overall distribution of microplastics by shape. The greatest concentration of plastic appears near the mouth of the Grand River, which connects to the major cities of Grand Rapids and Lansing. This explains the greater concentration of plastics in the sediment. Lake Erie was found to contain a significantly \begin{wrapfigure}{o}{3cm}
    \includegraphics[width=3cm]{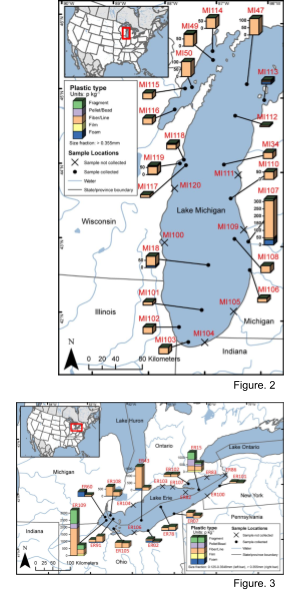}
        \caption{Spatial distribution of microplastics}

\end{wrapfigure} greater amount of plastic than Lake Michigan, with a maximum of over 3000 plastics per liter of water outside of Detroit, and a significant increase in fragment particles, which are commonly identified as tire wear particles (TWPs) and are associated with high levels of car usage. \cite{LenakerEtAlspatialdist}. 

    This data was used within the project primarily in order to determine the risks posed by benthic microplastics and thus decide what type of trawl would be created during the project. Although benthic plastics do pose a significant issue, it was determined that surficial plastics are both significantly easier to extract and just as (if not more) negatively efficacious on marine and littoral environments. In addition to this, surficial plastics are much more studied due to their ease of collection, meaning that modeling the effectiveness of a surficial trawl is more accurate than modeling the effectiveness of a benthic trawl due to having far more data points to consider when modeling.
    
\paragraph{Uniform Size Classification and Concentration Unit Terminology for Broad Application in the Chesapeake Bay Watershed}
  
    This paper establishes a common definition of what size of plastic debris constitutes microplastics, macroplastics, and nanoplastic. It was determined through investigation of several studies  that anything less than 5 centimeters constitutes a microplastic, and anything less than 1 micron was considered a nanoplastic. This paper synthesized the classifications of several previous studies in order to clarify the definitions of each aforementioned term \cite{TetraTechsize}.
    \begin{figure}[h!]
        \centering
        \includegraphics[width=8cm]{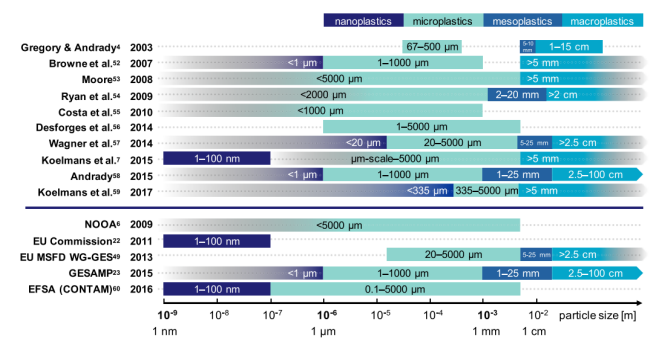} 
        \caption{Size classifications of microplastics}
        \label{fig:Size classifications of microplastics}
    \end{figure}
	This paper was used within this project for the primary purpose of establishing the classifications of and terminology used to describe microplastics in all of their forms. This project specifically discusses primarily micro and macroplastics, with brief mentions of nanoplastics interleaved throughout, and thus these definitions were used to ensure consistency in the terms used. In addition to this, the paper discusses the distribution of microplastics among size classifications, indicating that smaller microplastics are more numerous than larger, with a continual increase in quantity as size classification goes down. In this project, unfortunately, the net obtained had a somewhat large pore diameter of 300 microns, and as will be seen in the study Microplastic concentrations, size distribution, and polymer types in the surface waters of a Northern European lake, microplastic concentrations are much greater below 300 microns than above. Despite this disadvantage, 300 microns is the standard pore diameter of manta trawls used in many other studies referenced in this project for the purposes of modeling plastic collection, and thus using the same net specification likely increases the congruence between modeled effectiveness and real-world effectiveness.
 
\paragraph{Microplastic Contamination in Freshwater Environments: A Review, Focusing on Interactions with Sediments and Benthic Organisms}
	This study focused on compiling and analyzing data from previous studies on the concentration of microplastics within various rivers and other bodies of water. The study also discussed the different units of measurement used by the various studies on the subject and attempted to reconcile and compare some of them, as currently plastic quantities are measured by weight, particles per unit of area, particles per unit of volume, and total number of particles. Overall, this lack of standardization slows down progress in measuring microplastics and mitigating their negative effects, and harms the study of microplastics as a whole. In addition, this study examined the ecotoxicology of microplastics and their ways of accumulating in benthic areas \cite{BellasiBenthic}. 
	In this project, the data from this study was used primarily in the rationale, for explanations of the effects of microplastics on the environment, and for standardization of measurement for the modeling of the product’s effectiveness. Due to this paper’s illumination of the commonality of disparity in measurement systems across studies, special care was taken to ensure congruence in units of measurement throughout this study in order to ensure accuracy. In addition, this study helped with the explanation of microplastics’ effects on benthic sedimentary systems, where they can constitute up to 3\% of sediments by weight. Benthic organisms are disproportionately affected by microplastics due to their high consumption of high density plastics which fall into the benthic zone and become mixed with naturally occurring sediment to be consumed. In a continuation of this project, a method of removing microplastics from benthic sediment may be explored. 

\paragraph{Microplastic concentrations, size distribution, and polymer types in the surface waters of a Northern European lake}
    This study used a Manta Trawl along with a pump filter in order to measure the quantity of microplastics in Lake Kallavesi in Eastern Finland. The plastics were measured at 11 different sites around Kuopio, a moderately sized city with 118,000 inhabitants at the study’s time of writing. The average concentration of plastics found was 0.27 ±  0.18 microplastics/m3 of water when using a Manta Trawl, with the highest concentrations of plastics found was within the city’s harbor, with the lowest found underneath a road bridge in the city. When measured using pump filtration, however, the results changed significantly, with 1.8 ± 2.3 (>300 \textmu m), 12 ± 17 (100–300 \textmu m) and 155 ± 73 (20–100 \textmu m) microplastics/m3 found on average across the various sampling sites. This seems at first glance to indicate a significant failure of Manta Trawls to collect microplastics, but the authors of the study posit that this is instead a result of unreliability of pump filtration along with differing conditions at the time of measurement; the Manta Trawls were used in autumn and ran along a transect, whereas the pump filtration system was operated at one location, on a much smaller volume of water, in spring. Thus the authors say that the pump filtration data is likely much less accurate than the trawl data, and that more testing must be done to obtain unbiased results \cite{Uurasjarviconcentrations}.
    Within this project, the data from this study was used primarily to determine the best locations for the operation of the Manta Trawl and to aid in considerations of using a smaller pore diameter in future iterations of this project. In a future iteration, it is very likely that a smaller pore diameter would be used in order to collect more plastics, because as seen in the study’s  data, pore diameters around 20-100 microns may be able to collect hundreds of times more microplastics than pore diameters around 300 microns. In addition to that consideration, this study confirmed that areas closer to city drainage systems experience much higher concentrations of microplastics than those further away, although this data may not be entirely applicable due to the difference in scales and currents between Lake Kallavesi and the Great Lakes

\paragraph{Microplastic pollution in the surface waters of the Laurentian Great Lakes}
    This study collected data on the concentration of microplastics by location within the Laurentian Great Lakes Superior, Huron, and Erie. The authors collected this data by using a standard Manta Trawl, and they found that of the lakes studied, Lake Erie was by far the most polluted by area, with concentrations of  microplastics ranging from 4,686 microplastics / km2 to an immense 466,305 microplastics / km2. These plastics were all found at the surface, and measurements were based upon area instead of volumes, meaning that these figures are likely significant underestimations due to the often significant quantity of semi-buoyant or non-buoyant plastics within the Great Lakes, especially given that the hydrologically connected St. Lawrence River has been found to contain high concentrations of benthic microplastics in separate studies, indicating that many of the lake’s plastics are benthic \cite{Eriksenlaurentian}.
    This study was used primarily for the rationale of the project, especially when considering the feasibility of the given solution and the resources which would be necessary to enact it as a long term solution to the problem of microplastics. Given the variety of sampling locations, it is possible to calculate the rough number of Manta Trawls needed to clear large sections of lakewater of microplastics. Based on data found in this study, a fleet of 802 self-sufficient Manta Trawls operating every day for roughly 15 years would remove over 95\% of microplastics in Lake Erie, taking into account lowering concentration of plastics and the yearly deposition of more plastics into the lake. Although a project like this would require significant investment (roughly \$800,000 in material costs), these calculations show that (at least in theory), solutions to the problem of microplastics can be found, and the problem is not a hopeless one. Despite this, it may very well be far more feasible to scale a Manta Trawl up significantly and use boats to propel it in order to more efficiently solve the problem, however that scenario is much harder to model due to non-existence of data using such trawls. More information as to how the above numbers were calculated is found in the Modelling section of this paper.

\paragraph{Influence of wastewater treatment plant discharges on microplastic concentration in surface water}
	One significant source of microplastics in the waterways of highly developed countries is output from wastewater treatment plants (WWTPs). WWTPs are very effective at removing macroplastics, harmful chemicals, and other pollutants, however they lack the same efficiency in removing microplastics from wastewater. This leads to significant output of microplastics from WWTPs which can have significant negative effects on aquatic ecosystems downstream from the plants. As seen in the charts above, areas downstream of WWTPs can have upwards of 300\% the quantity of microplastics when compared to areas just upstream of the plants \cite{EstahbanatiWaste}. 
 \begin{figure}[h!]
     \centering
     \label{fig:WWTP graph}
     \includegraphics[width=8cm]{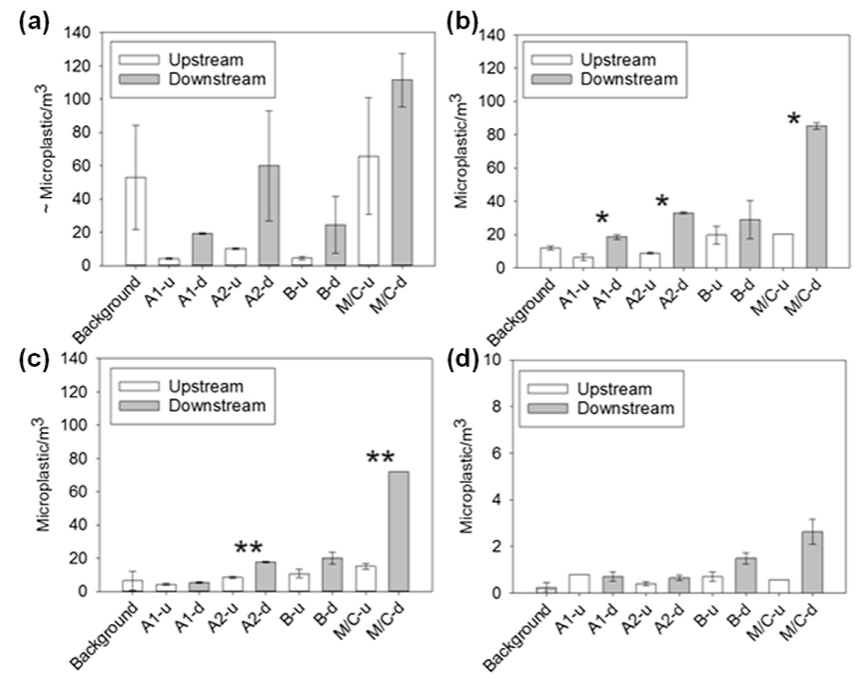} 
     \caption{Wastewater treatment plants' microplastic output}
      \end{figure}

	This information was crucial in understanding the greatest contributors to microplastics pollution in rivers. Past this understanding, the information here is also particularly relevant in considering the potential sources of microplastics in the Milwaukee River where testing was performed. In this project’s testing, abnormally low concentrations of fibers were collected when compared to other studies’ findings, which could be explained by the lack of WWTPs upstream of the testing location. WWTPs typically produce mostly secondary microplastics, and most secondary microplastics are fibers or lines produced by appliances like washing machines when washing polyester or other plastic clothes. WWTPs also produce a significant amount of primary microplastics, leading to great environmental damage.

\paragraph{The influence of exposure and physiology on microplastic ingestion by the freshwater fish Rutilus rutilus (roach) in the River Thames, UK}
    This study examined the ability of microplastics to be consumed by the freshwater fish Rutilus rutilus (common roach). This fish is a crucial part of ecosystems across Europe and Western Asia, constituting the greatest biomass of any fish in several environments, thus being of critical importance in the maintenance of food webs throughout the world. This study examined microplastic uptake by these fish, finding that in the River Thames by catching and dissecting 64 fish from various locations between Oxford and London. After examination, it was found that nearly 1/3 (31.8\%) of these fish had consumed one or more microplastic, with the percent having consumed plastics increasing as samples were taken further and further downstream. This indicated that microplastic concentrations were increasing downstream, as urban centers deposited buoyant microplastics onto the river \cite{HortonRutilus}.
    In this project, this study aided in developing the rationale and furthered the understanding of microplastic sources and rates of uptake by aquatic organisms. By focusing on a type of fish which is very common, the study was able to illustrate the scale of microplastic pollution throughout waterways. It was made clear that this issue is incredibly widespread, and that consumption of microplastics was common among these fish. Another aspect of this study which was useful to this project was the quantification of microplastic concentrations at many locations on the same river. This allowed for a clear illustration of the sources of plastic pollution (primarily population centers and industrial sites) by examining the quantities of plastic found in fish at several sites downstream of each other. 

\subsection{Construction \& Operation}
\subsubsection{Manta Trawl}
\paragraph{Manta Trawl Trawling Protocols}
	This document was prepared by the 5 Gyres Institute, an organization focussed around environmental preservation through the monitoring and removal of plastics from the world’s lakes and oceans, specifically focussed around microplastics. This document outlines the different protocols which researchers should follow when collecting data using Manta Trawls, with provisions for acceptable wind speeds, trawling speeds, and trawl location. Specifically, the document states that trawling should be done only when water is observed to be less than a 4 on the Beaufort scale, trawling should be performed under 3 knots, and the trawl should be deployed either on the port or starboard sides of the ship, but not at the stern or aft in order to prevent data from being affected by the boat’s wake. (5 Gyres Institute, 2018)
	This document was used primarily to determine both the ideal operating conditions for an autonomous version of the trawl and the speed at which such a trawl must travel in order to collect microplastics at a quantity greater than or equal to manually pulled trawls. This speed was determined to be roughly 2 knots by the study, and this was taken into consideration when constructing additional parts of the trawl –both in that relatively powerful motors must be purchased to propel a trawl at at least 2 knots(or more when considering currents or wind) and in that too much consideration should not be lent to the hydrodynamics of the trawl or of constructed devices, as drag matters far less at such low speeds.
 
\subsubsection{New Constructions}
\paragraph{How to Use an Ultrasonic Sensor}
	This article written by MaxBotix goes over how to use ultrasonic sensors in great detail. The article includes which terminals on the sensor to solder wires to on the sensors as well as very comprehensive instructions on how to program the sensors using pulse width modulation and analog voltage. In addition to the instructions on how to connect physically to the Arduino, it also contains the code to run the sensors itself, creating an altogether very useful resource for programming and connecting sensors to the Arduino Mega 2560 used in this project. The article also has tutorials on basic programming concepts which are a great starting point for any beginning experienced programmer just getting started with Arduinos and ultrasonic sensors. (MaxBotix, 2021)
	In this project, this tutorial was used for a significant amount of the base ultrasonic programming work, and although it wasn’t used for the more complicated and unique parts of the code, such as moving averages, motor control through the motors’ ESCs(Electronic Speed Controllers), or sensor output stabilization, it helped with some of the more basic functions. Overall, it significantly improved sensor performance and lowered programming time as a whole by providing tried and tested solutions for managing the project’s ultrasonic sensor. 

\paragraph{NewPing Arduino Library}
    The newPing library is a library that provides a host of methods and functions, allowing the easy use of ultrasonic sensors with Arduinos. This library’s methods also provide optimization for many already existing functions built into the default Arduino architecture. In addition it features automatic conversion from the microseconds outputted by ultrasonic sensors to centimeters or inches that can be more easily understood by humans. Overall, the library greatly simplifies communication and interaction between the Arduino and ultrasonic sensors. (Eckel, 2022)
    In this project, the NewPing library was used for its single-pin connection capabilities. Single-pin connection allows an Arduino to connect with an ultrasonic sensor using only a single connection wire instead of two separate wires (one for triggering the ultrasonic sensor and one for receiving the signal). In addition the library allows for easy conversion of ultrasonic sensor data from microseconds between pulse pings and centimeters or inches, greatly simplifying the code to create moving averages of data values. These moving averages were used to stabilize the data received by the sensors and to lessen the impact of outliers caused by sensor inconsistency. Overall, NewPing allows the code to run significantly faster and more efficiently, with fewer actual lines of code and a simplification of the code overall.
	
\paragraph{3D Printing Watertight and Air Tight Containers}
	This article goes over techniques and settings that can be used to waterproof 3d prints and ensure that components inside are not damaged either by liquid or by growth of biological materials within the print. Typical 3d printed parts have very thin walls which are supported by internal lattices to create a structure that uses very little material while remaining structurally sound. This works very well for most applications, however it causes issues when attempting to prevent fluids from entering a print, as minor damage or microscopic holes within the print can lead to flooding and further damage. To prevent this, the article suggests changing 3d printer settings to ensure a wall thickness greater than 0.068 inches to ensure a sufficiently thick seal, in addition to using multi-layered walls. (Instructables, 2018)
	In this project, the information within the article was used to ensure that the vessel 3d-printed to contain the electronics on board the trawl could protect the devices within from any water damage. It is likely that the bottom section of the 3d printed vessel will be underwater or contacted by water for the majority of the trawl’s active time, so it is critical that water cannot contact the electronics inside. Because of this, the 3d printed containers were printed using line infill instead of lattice infill, and infill percentages were relatively high; 35\% for the container as opposed to the 25\% used for the clamps. 

\section{Engineering Goal}
    The goal of the project is to develop an autonomous submersible based upon a Manta Trawl which is capable of collecting microplastics in waterways through filtration in order to lower their quantity and thus increase the health of organisms living near and inside of those waterways. This, ideally, could serve as a model to scale up either the quantity of or scale of microplastic-collecting apparatuses within the Great Lakes and other bodies of water, which is currently nearly non-existent. To be considered successful, this project must fulfill two goals; first, it must show that microplastics could be significantly lowered in quantity through widespread use of a filtration system, and two, must show that a filtration system could be automated and deployed throughout an area by creating a prototype of such a system and testing it.

\section{Procedure}
\subsection{Overview}
    The final product of this project is based upon the Manta Trawl system which is currently used by researchers in many of the studies cited in order to measure quantities of microplastics in aquatic ecosystems. The primary modifications will be the addition of electric motors with propellers in order to give the final product its own propulsion system, along with an autonomous guidance system run by an Arduino Mega2560 along with an ultrasonic sensor which will allow it to successfully navigate whichever body of water it is in. Thus the construction of the device can be divided into two primary parts; first is the building of the additions’ housing and hardware, and second is the development of software systems for the trawl. The procedure will thus be split as such. 
\subsection{The Initial Trawl}
    The initial Manta Trawl was received from Mr. Peter L. Lenaker of USGS by agreed delivery to Nicolet High School after contact via email on 11/10/22(mm/dd/yy). There is one major concern regarding the size of the trawl, which is the quantity of microplastics which the trawl could hold without beginning to spill more than it collects. To calculate this quantity, we take the following steps.
\begin{enumerate}
    \item Calculate the total volume of the trawl
    \begin{enumerate}

    \item The trawl’s shape is a rectangular pyramid, so we can calculate that: 
            \subitem \({V=\frac{L*W*H}{3}}\) 
            \subitem \({0.25m^3=\frac{1.5*1*0.5}{3}}\) 
    \end{enumerate}
    \item This result is converted from \({m^3}\) to \({mm^3}\) for ease of calculation on the scale of microplastics. This turns \({0.25 m^3}\) to \({250,000,000 mm^3}\).
    \item Assuming the worst case scenario, all of the microplastics collected are the maximum possible size of a microplastic, i.e. 5 mm diameter sphere
    \item Calculate quantity of microplastics which could fit into the trawl
    \begin{enumerate}
        \item \({V_{microplastic}=\frac{4}{3}\pi r^3=\frac{4}{3}\pi (2.5)3=65.5mm^3}\)
    \end{enumerate}
    \begin{enumerate}
        \item \({250000000/65.5 \approx 3,816,793}\) microplastics
    \end{enumerate}
\end{enumerate}
    The rate at which the trawl fills up will be incredibly dependent on where it is located within the Great Lakes, as concentrations of microplastics vary over a range from roughly 0.001 mps/m\textsuperscript{3} up to 0.932 mps/m\textsuperscript{3} in the most heavily polluted areas across all 5 lakes. On average in Lake Erie (the most polluted of the Great Lakes), a trawl will collect roughly 220 mps/hour (0.104 mps/m\textsuperscript{3}) based on current concentrations, meaning that the trawl would take nearly two years to fill up, whereas the most polluted areas may see a full trawl within 3 months –although both of these scenarios make the unrealistic assumption that concentrations of microplastics remain the same throughout the duration of the collection. Overall, this means that plastic collection in the most heavily polluted areas will need to be much more heavily monitored in order to ensure continued operation of these devices. It must also be noted that these times may be over or under estimations due to a few key factors, those primarily being speed and organic matter. The total volume of microorganisms, algae, and other non-plastic debris which will enter the trawl is a nearly incalculable value, especially given the possibility of regional phenomena like algae blooms adding a high degree of uncertainty to the calculations. In the samples collected in the Milwaukee river at Estabrook Park, this material constituted a very significant amount of the total volume, with visual estimation placing it somewhere between 50\%-75\% percent of the overall volume. This is a massive number, however it is likely to be completely unrepresentative of the lake environments where the trawl will primarily operate for two reasons:
\begin{enumerate}
    \item The Milwaukee river has very high quantities of algae due to being warm, slow-moving, and shallow at the location of testing.
    \item The draught of the trawl ( \(\approx\) 50 cm) is very similar to the depth of the river (\(\approx\) 60 cm), meaning that benthic microorganisms were almost certainly present in the sample.
\end{enumerate}
    These factors mean that the ratio of organic matter to microplastic debris is likely overestimated, and samples from a lake will likely be far different from those in a small and shallow river. Despite this, collection of non-plastic material will certainly hasten trawl filling, however, other factors counterbalance this to an unknown amount. One such factor is that of speed, as a full trawl would have much greater mass than an empty one. A full trawl would not weigh more, as the plastics inside would be buoyant, however the trawl motors would struggle greatly to pull the greater mass of the plastics, so as more material is collected the rate of collection would increase and spillage from currents pulling plastics out of the trawl may increase.
    
\subsection{Building the Housing and Propulsion Systems}
    The housing for the primary components began with a CAD file created in Solidworks, an industry standard for engineering and design, and after being designed the file was 3D printed on Nicolet High School's Dremel 3D printers. It will be mounted to the trawl with 3D-Printed clamps in an effort to preserve the structure of the Manta Trawl and preclude any damage to the structural integrity of the frame. Once 3D printed, the housing will be filled with the Arduino Mega2560, ultrasonic sensor, 2 batteries(one 12-volt for the motors and one 9-volt for the Arduino Mega), and all other components which need to be protected from water such as the GPS and ultrasonic sensor control module. After the attachment of the housing, the two motors mentioned in the materials list will be attached to the side of the trawl using the same clamp system used to attach the electronics’ housing with a slightly modified section for mounting the motors using 4 M4 specification screws each. After attaching the motors, they will be connected to the Arduino for control, and the front of the trawl will be covered with a larger pore diameter mesh to prevent the capture of macroorgansims like fish or macroplastics like bags within the system while retaining the capability to collect plastics.

\subsection{Creating the Guidance Systems}
    To create the guidance system, an Arduino Mega 2560 will be used as the computer along with one ultrasonic sensor for use in collision detection and one GPS transponder, as the system will need very little guidance for the majority of its operating time. The ultrasonic sensor will solely serve the purpose of preventing collisions with the banks of the body of water which the trawl is operating in or with any other vessels within the water.

\subsection{List of Control Box Materials}

\begin{enumerate}
    \item 1 - Arduino Mega 2560
    \item XALXMAW WAGO wire connectors
    \item 9v rechargeable battery
    \item 1 - Zeee 15,000mAh 11.1V Lipo Battery with EC5 connector
    \item XT60 and EC5 Wire Connectors
    \item Male-female, male-male, and female-female dupont cables 
\end{enumerate}

\subsection{Construction and CAD}
\begin{figure}[h!]
    \centering
    \label{fig:fullPic}
    \includegraphics[width=8cm]{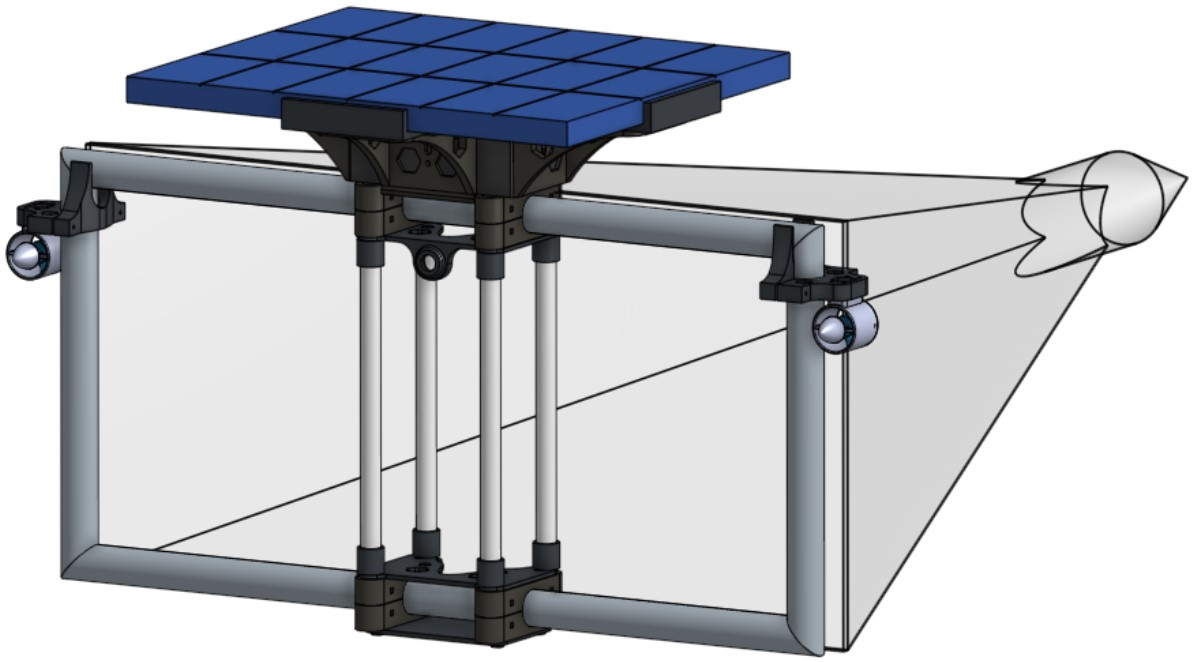} 
    \caption{Full 3D model}
\end{figure}

\subsubsection{Construction Materials List}
    The model above shows the complete assembly of the trawl, with every part represented in an accurate position on the trawl’s frame. There are 15 distinct 3D printed parts in total, with 47 instances of those parts cumulatively. The 15 parts are: 
\begin{enumerate}
    \item Trawl attachment clamp (8)
    \item Underwater sensor module (1)
    \item Control box (1)
    \item Side wing (2)
    \item Stem-Stern wing (2)
    \item Motorside motor clamp (2)
    \item Trawlside motor clamp (2)
    \item Trawl attachment aligner (1)
    \item Bottom holding plate (1)
    \item Wire cap (2)
    \item Trawl attachment pin (8)
    \item Lid (1)
    \item Front solar panel support (2)
    \item Side solar panel support (2)
    \item PVC Pipe adapter (8)
\end{enumerate}
    In addition to these, there are 6 distinct purchased or otherwise non-manufactured parts outside of the control box. These are: 
\begin{enumerate}
    \item Neuston net/manta trawl (1)
    \item ¾ in x 1.2ft PVC pipe (4)
    \item Diamond Dynamics underwater thruster (2)
    \item Renogy 50W 12V Monocrystalline solar panel (1)
    \item X-Haibei Marine Mooring buoy (2)
    \item Assorted bolts and nuts (12)
\end{enumerate}
\subsubsection{Construction Steps }
	The construction of the device begins with the central piece, the control box. This part contains all of the electronic components of the device except for the sonar and motors, and it is the most important part of the trawl –thus most of the design’s elements are focused around its stability and continued functioning. The technical drawing of the control box is shown below, note that this is a very complex design and some dimensions are excluded from the drawing for the sake of clarity. None of the excluded dimensions are critical to the design of the box, and they are primarily aesthetic considerations such as filets.
 \begin{figure}[h!]
    \centering
    \label{fig:boxTech}
    \includegraphics[width=8cm]{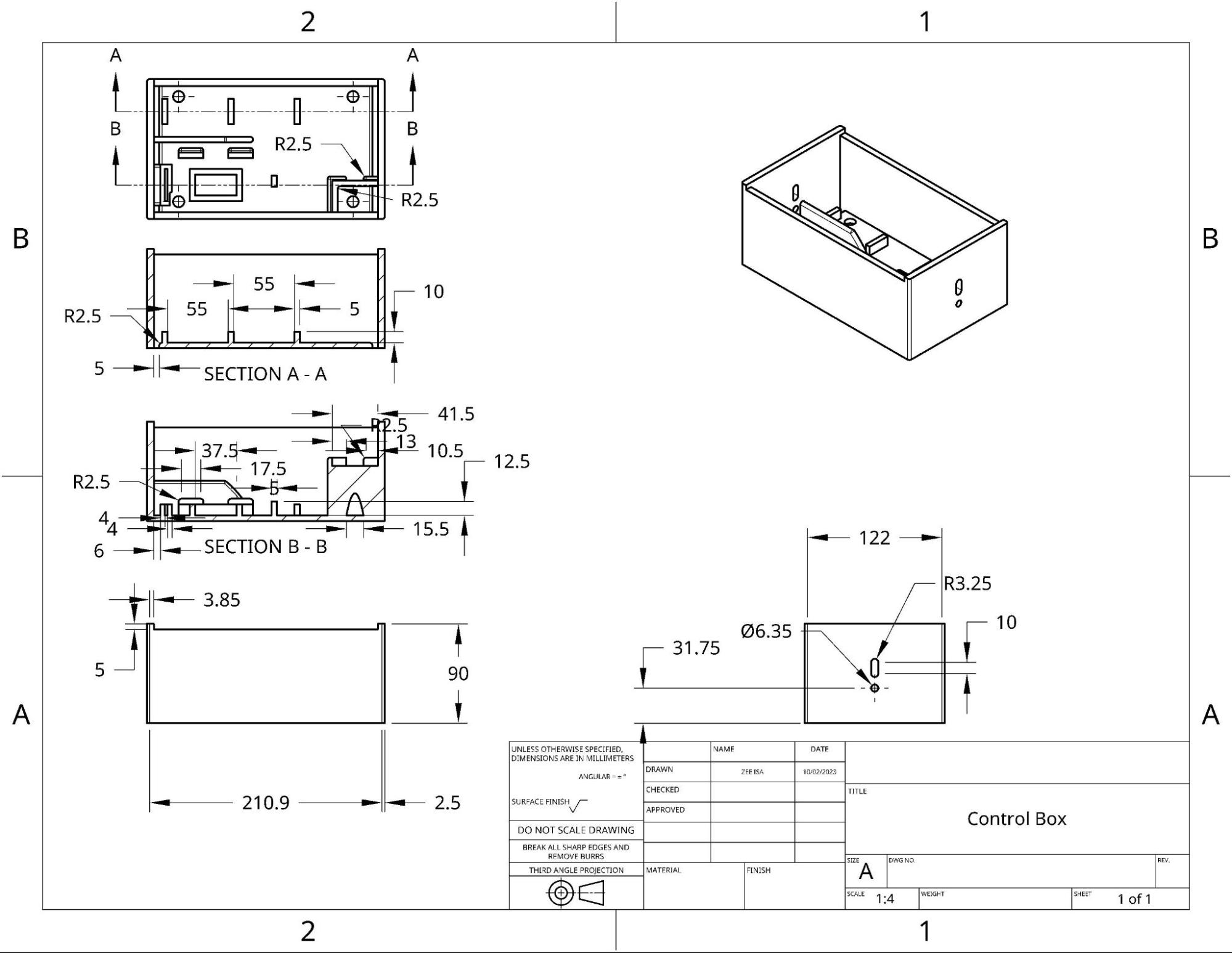} 
    \caption{Control box technical drawing}
\end{figure}

Connected to the control box via 2, ¼”\(\varnothing\) x ¾” bolts on either side are the side wings. These pieces are incredibly important to the longevity of the trawl, as they serve as the primary supports of the solar panels mounted on the trawl. The side wing’s technical drawing is shown below, along with an image showing how the two wings attach to the control box.
\begin{figure}[h!]
    \centering
    \label{fig:SideWingTech}
    \includegraphics[width=8cm]{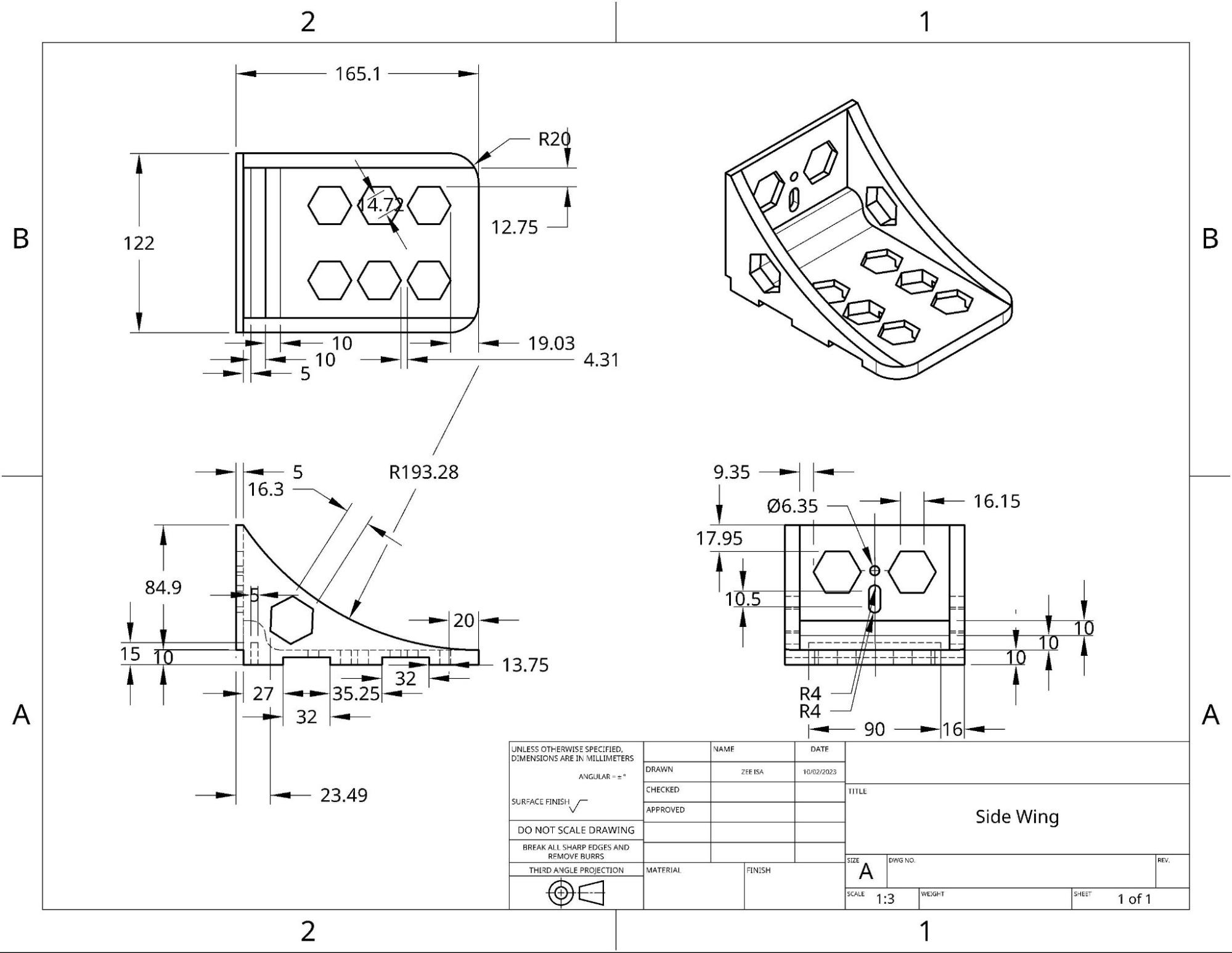} 
    \caption{Side wing technical drawing}
\end{figure}

\begin{figure}[h!]
    \centering
    \label{fig:BoxWWingsPic}
    \includegraphics[width=8cm]{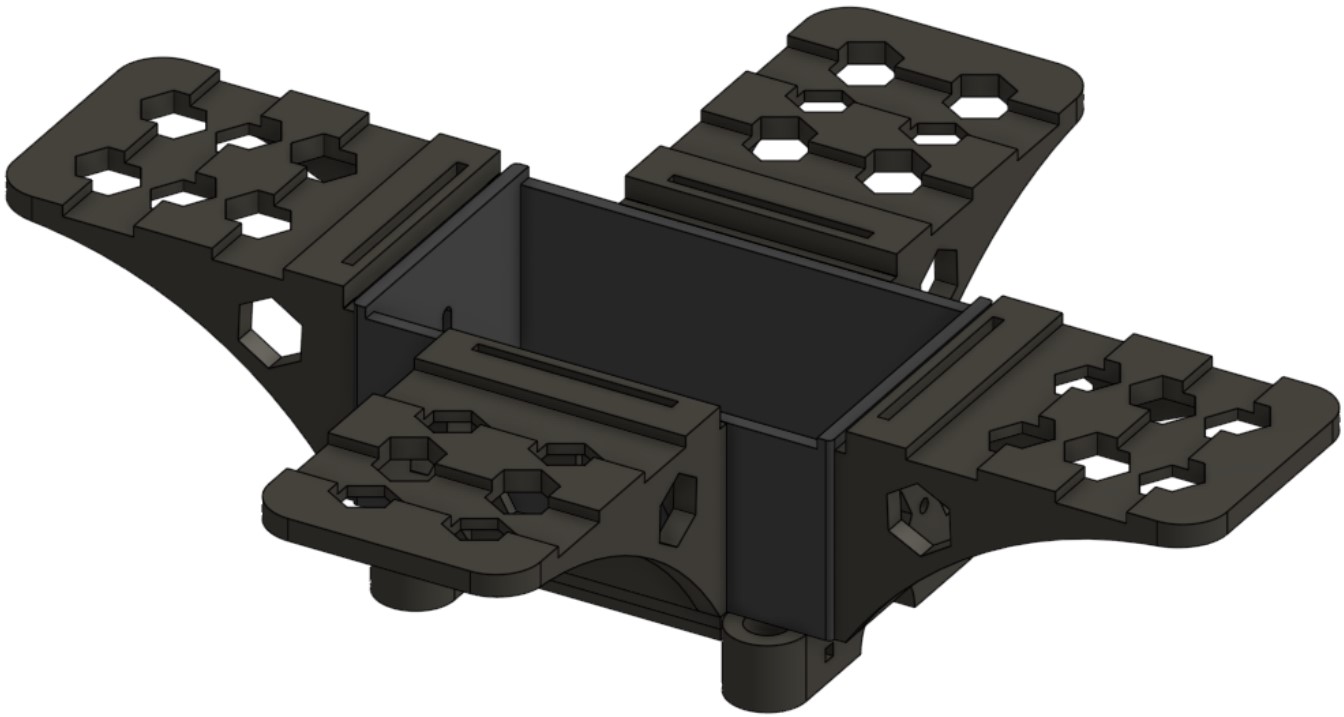} 
    \caption{Side wing 3D model}
\end{figure}

In addition to the side wings, there are similar wings mounted at the stem and stern of the control box to provide greater support for the solar panel. These wings have only one modification, that being a small alignment extension allowing them to be more easily aligned and attached to the control box using 2-part epoxy. The epoxy in this case, though not quite as strong as the screws holding on the side wings, are sufficiently strong that it can completely hold up the solar panels. These stem and stern wings are not bearing as much vertical load as the side wings due to the greater rigidity of the side wings’ connection (bolt versus epoxy), and their only purpose is to prevent the solar panel from sliding backwards and forwards during operation. They also prevent any upward forces from dislodging the solar panel due to the panel being screwed into the wings’ extenders. These extenders are the pieces which transfer load from the panel into the side wings, and they were necessary due to the unfortunately limited build volume of standard 3D printers. They attach to the stem, stern and side wings through their hexagonal holes and use 2-part epoxy to hold the pieces together. Below are shown technical drawings of these pieces and how they connect together. 

\begin{figure}[h!]
    \centering
    \label{fig:SidePanelMountTech}
    \includegraphics[width=8cm]{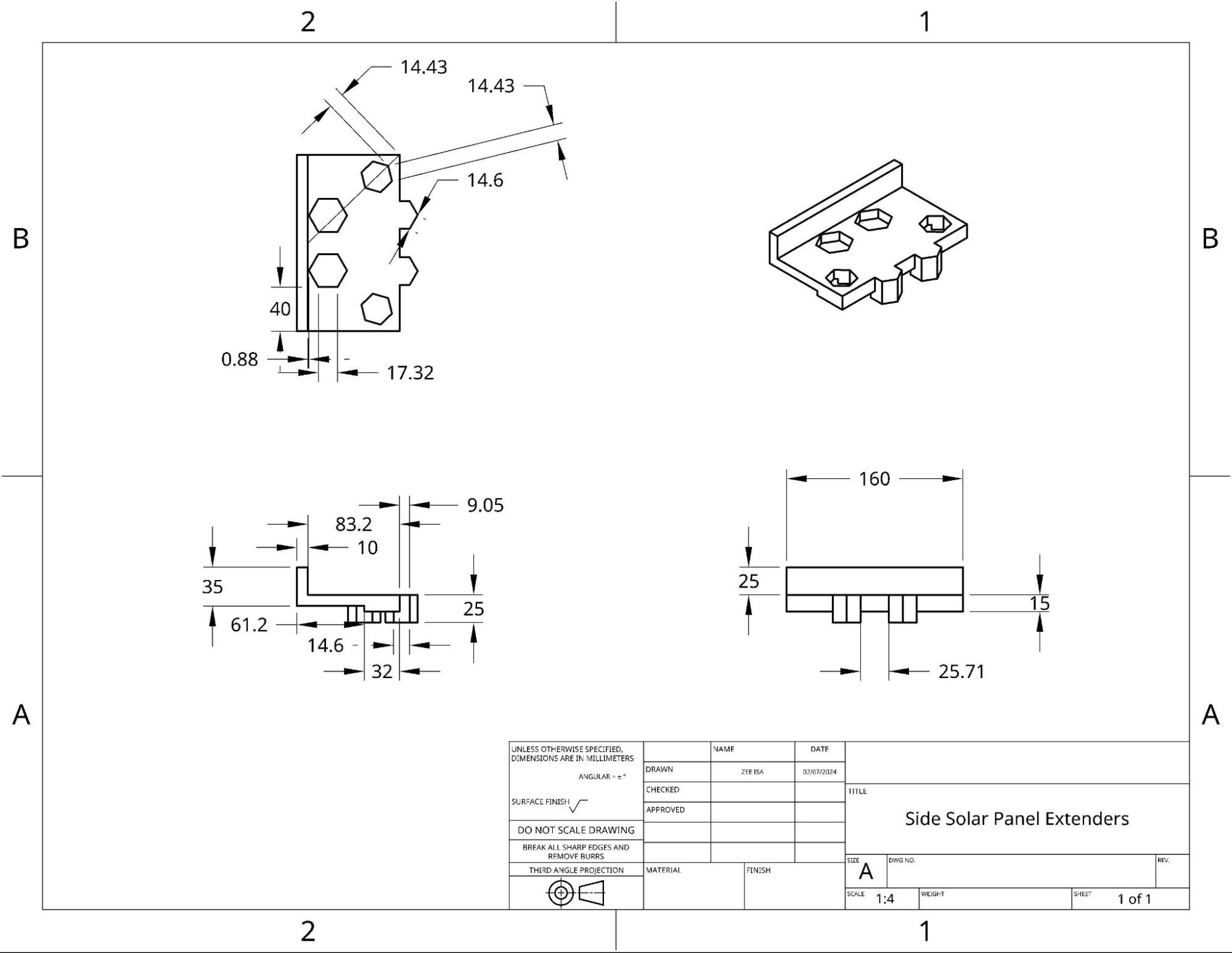}
    \caption{Side solar panel mount technical drawing}
\end{figure}

\begin{figure}[h!] 
    \centering 
    \includegraphics[width=8cm]{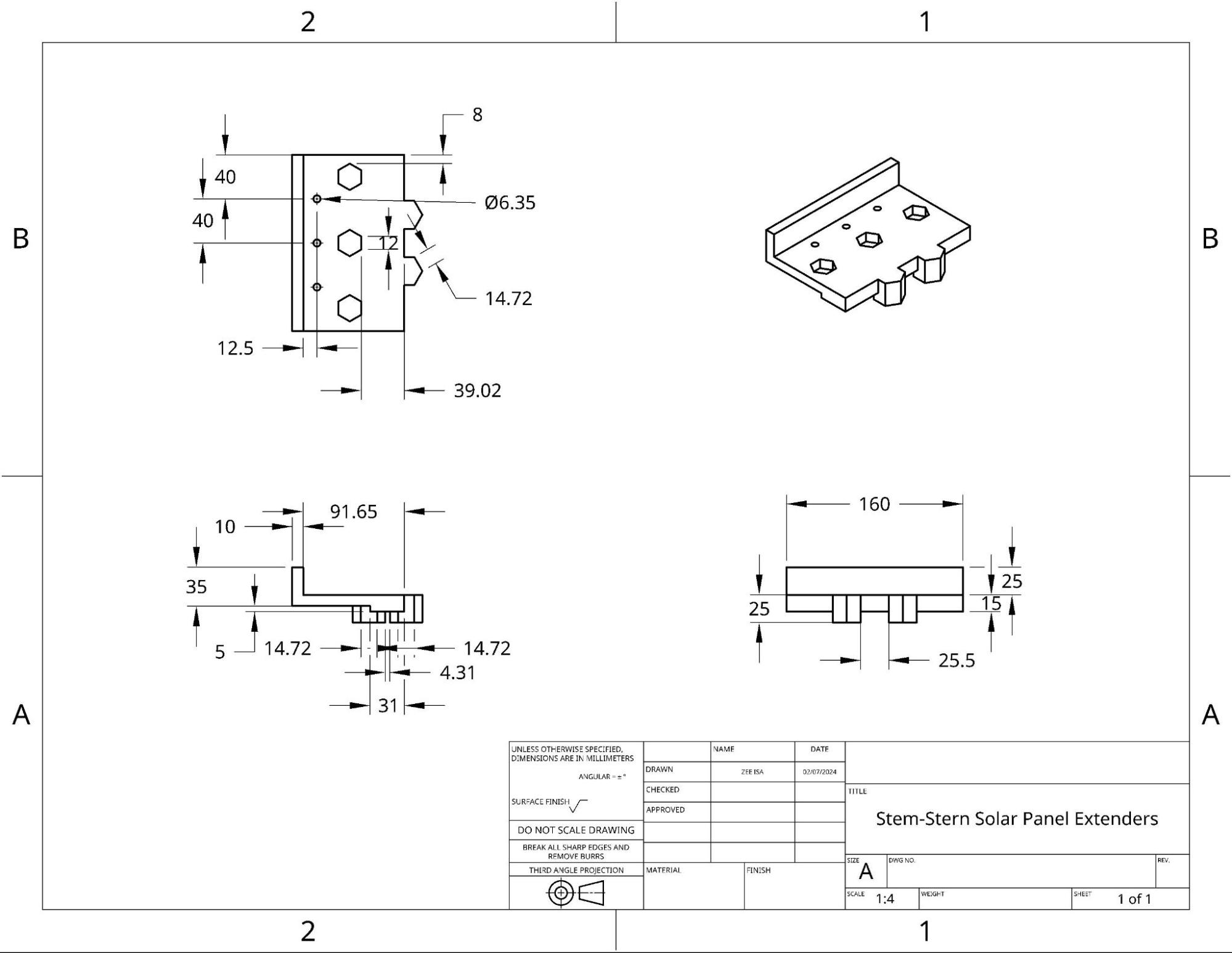} 
    \caption{Front and back solar panel mount technical drawing}
\end{figure}
One more of the most important aspects of the design and a major area of improvement are the trawl attachment pieces. These are standardized pieces which hold the control box to the trawl, and they also act significantly in stabilizing the box and holding the underwater ultrasonic sensor. The stabilization which these pieces provide comes by using PVC pipes to connect the top and bottom sections of the trawl together. The technical drawing for this piece is shown below, along with a model of the connections it uses to stabilize the control box.

\begin{figure}[h!] 
\centering 
\includegraphics[width=8cm]{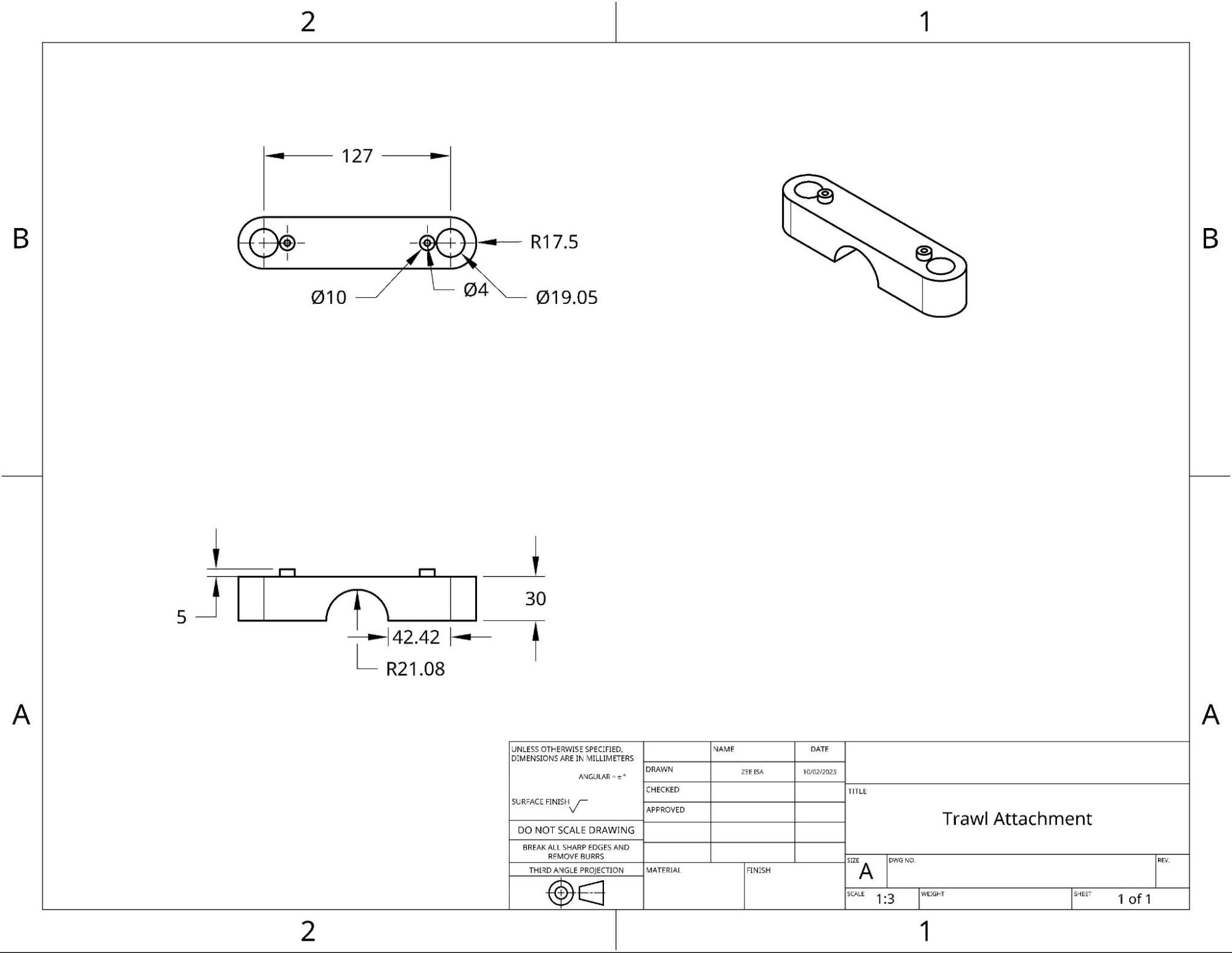}
\caption{Trawl attachment technical drawing}
\end{figure}

To hold these attachment pieces together, pins are used which can pressure fit into the top of the attachment pieces and hold down any plates attached to the pieces. The technical drawings for this pin and and the attachment piece are shown below. 

\begin{figure}[h!] 
\centering 
\includegraphics[width=8cm]{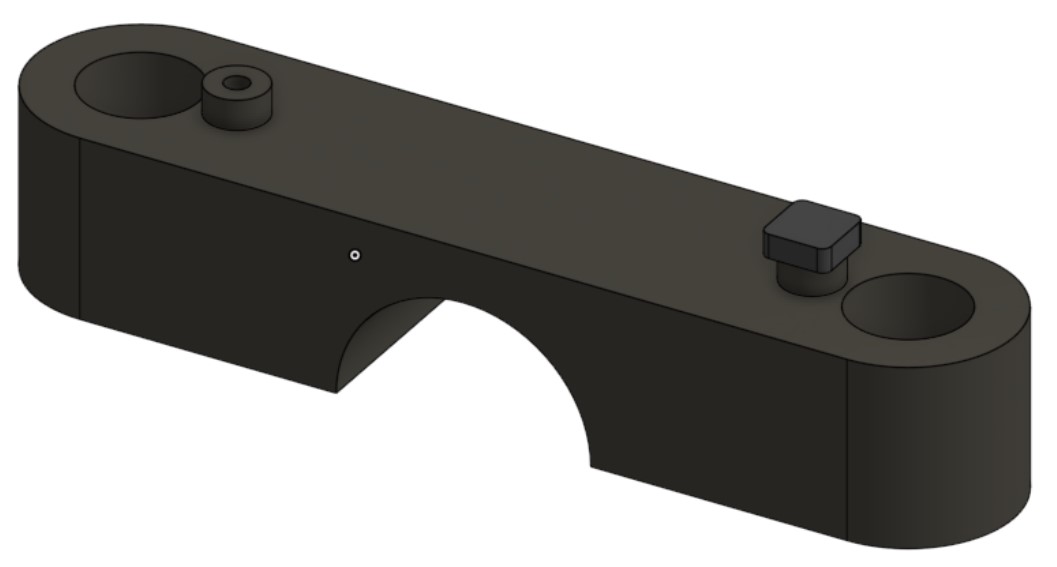}
\caption{Trawl attachment picture}
\end{figure}

\begin{figure}[h!] 
\centering 
\includegraphics[width=8cm]{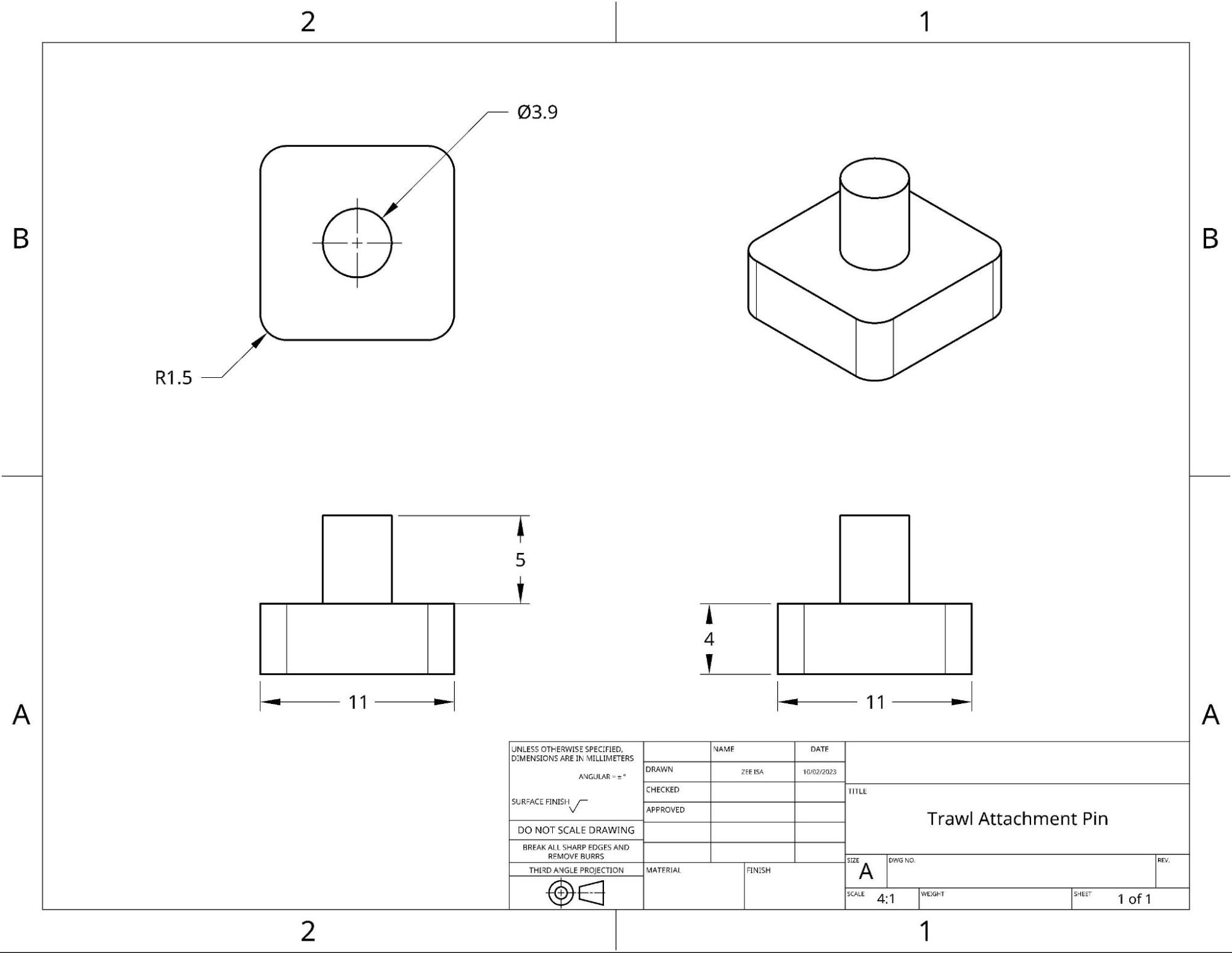}
\caption{Trawl attachment pin technical drawing}
\end{figure}

The underwater sensor module holds the ultrasonic sensor of the craft, in a significant shift from the previous trawl’s above-water placement of the sensor. This change allows the sensor to sense below water-obstacles much more effectively, and allows the trawl to steer away from shallow water and beaches far more effectively than an above-water sensor can. The technical drawing for this piece is shown to the side. In addition to this piece, there is also an alignment plate on the bottom side of the trawl’s lower bar which keeps the PVC pipes in alignment with the top of the trawl. This helps maintain the stability of the control box even more, and prevents mechanical forces experienced by the trawl from shifting the pipes to either side. A model showing both the ultrasonic sensor module and bottom alignment plate is shown below, along with the technical drawing of the bottom alignment plate. This primary structure at the center of the trawl is finished with the lid, which holds up the solar panel and forms a water-resistant and easily resealable top for the control box. The lid has one primary sealing piece with a minimal profile in order to allow the solar panels’ wires to bend sideways and come into the main control box without imparting significant stress onto the wires. 

\begin{figure}[h!] 
\centering 
\includegraphics[width=8cm]{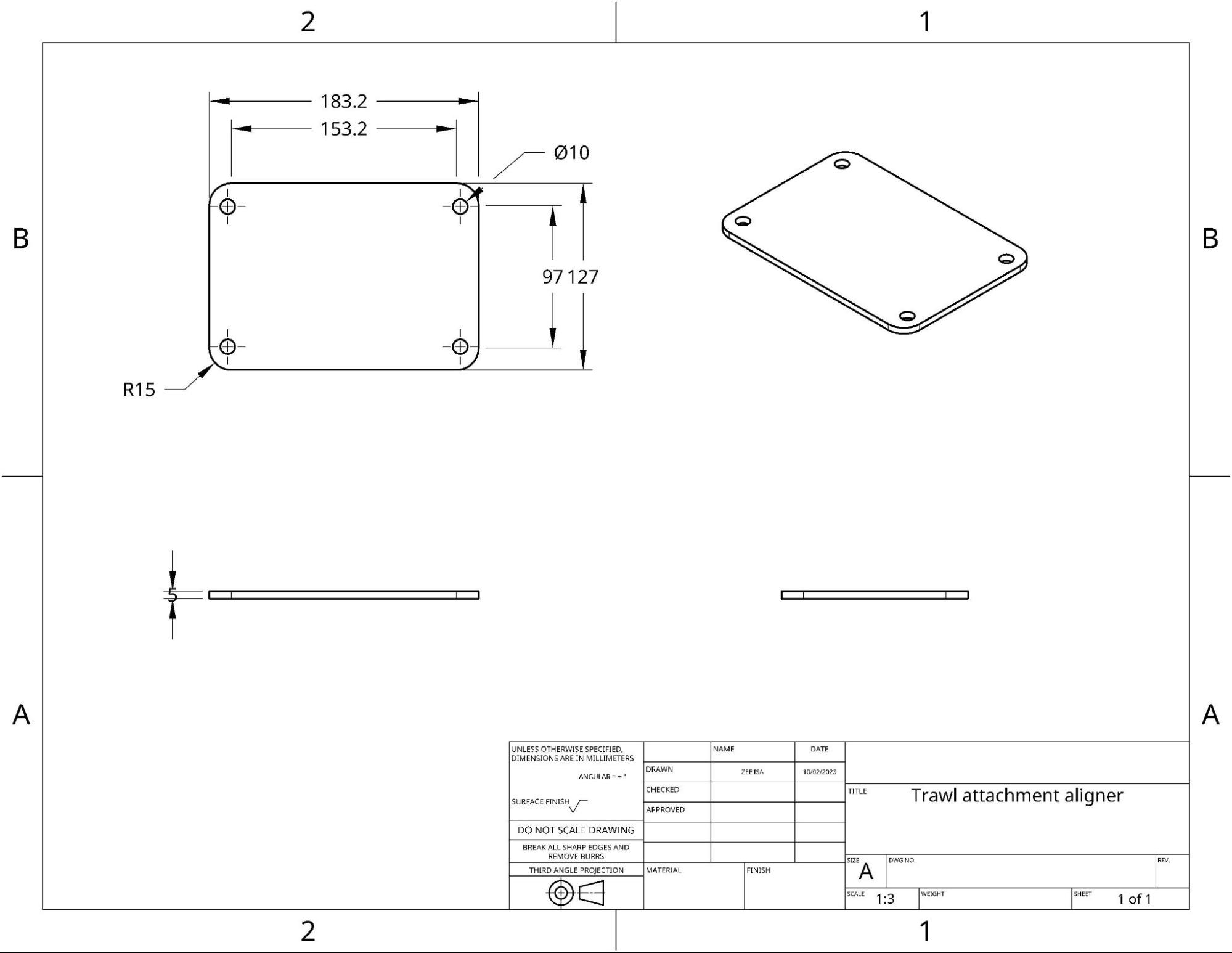}
\caption{Bottom plate technical drawing}
\end{figure}

\begin{figure}[h!]
\centering 
\includegraphics[width=8cm]{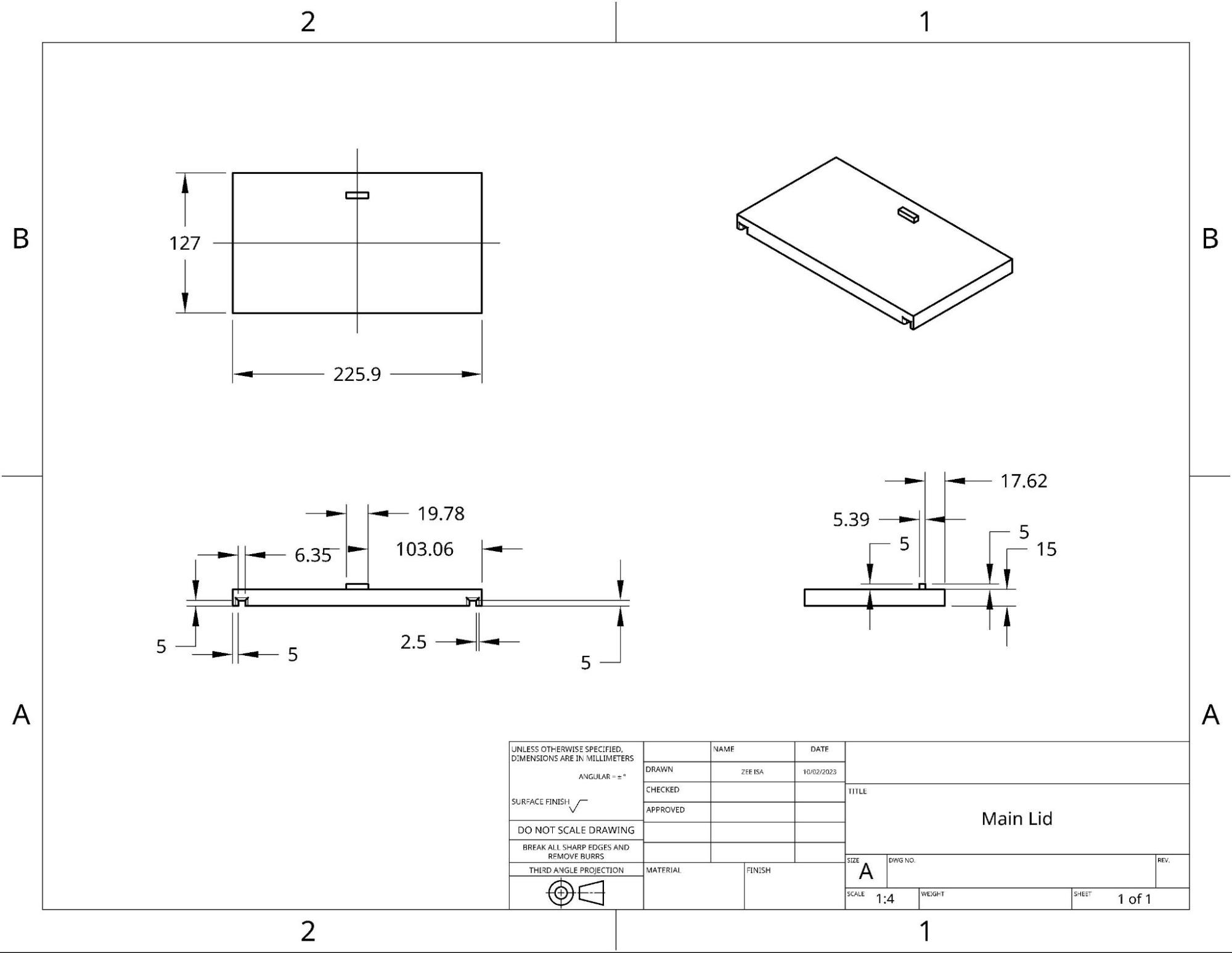}
\caption{Lid technical drawing}
\end{figure}

\begin{figure}[h!] 
\centering 
\includegraphics[width=8cm]{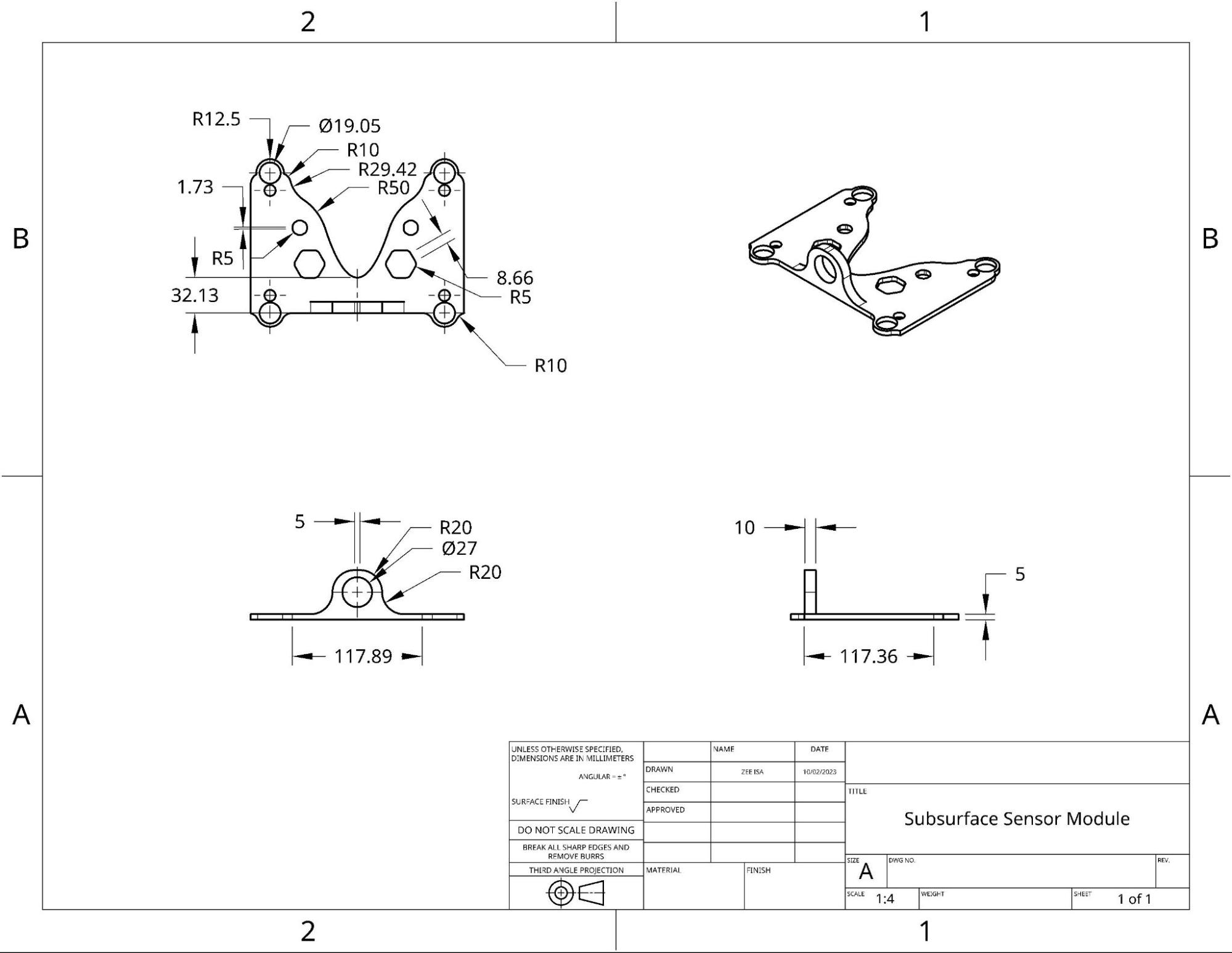}
\caption{Sonar sensor mount technical drawing}
\end{figure}

The lid locks into place through a pressure fit and is covered by a solar panel to ensure a strong fit and to make the control box assembly simultaneously sturdy and very difficult to take apart with the semi-random mechanical forces of wind, waves, or even collisions while making it very easy to take apart manually for maintenance and modification. By using bolts to hold the solar panel onto the top of the control box through the stem and stern wings, the device can also be very easily sealed. The last pieces of the control box are the two wire caps on either side of the box which prevent water from entering the box where the wires from the motors enter the control box. The technical drawing of these caps is shown below.

\begin{figure}[h!] 
\centering 
\includegraphics[width=8cm]{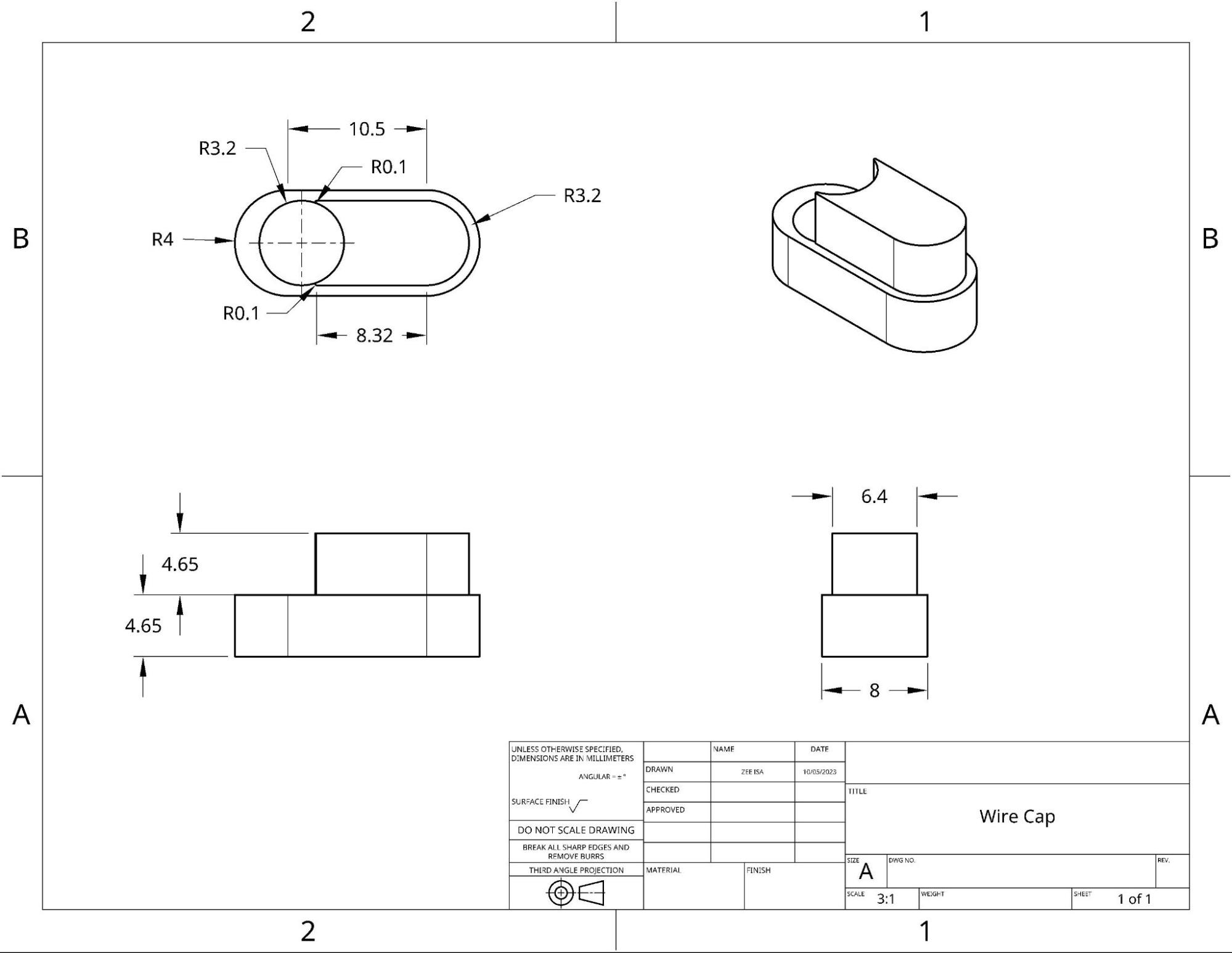}
\caption{Wire cap technical drawing}
\end{figure}

In addition to the aforementioned bottom alignment piece, a second similar piece modeled after the ultrasonic sensor plate holds the trawl attachment pieces onto the top of the bottom bar of the frame. This piece along with the ultrasonic sensor plate are held apart from each other by 4 PVC pipes which are subsequently connected to the trawl attachment pieces through 8 pipe adapter pieces, the technical drawings for which are shown below. 

\begin{figure}[h!] 
\centering 
\includegraphics[width=8cm]{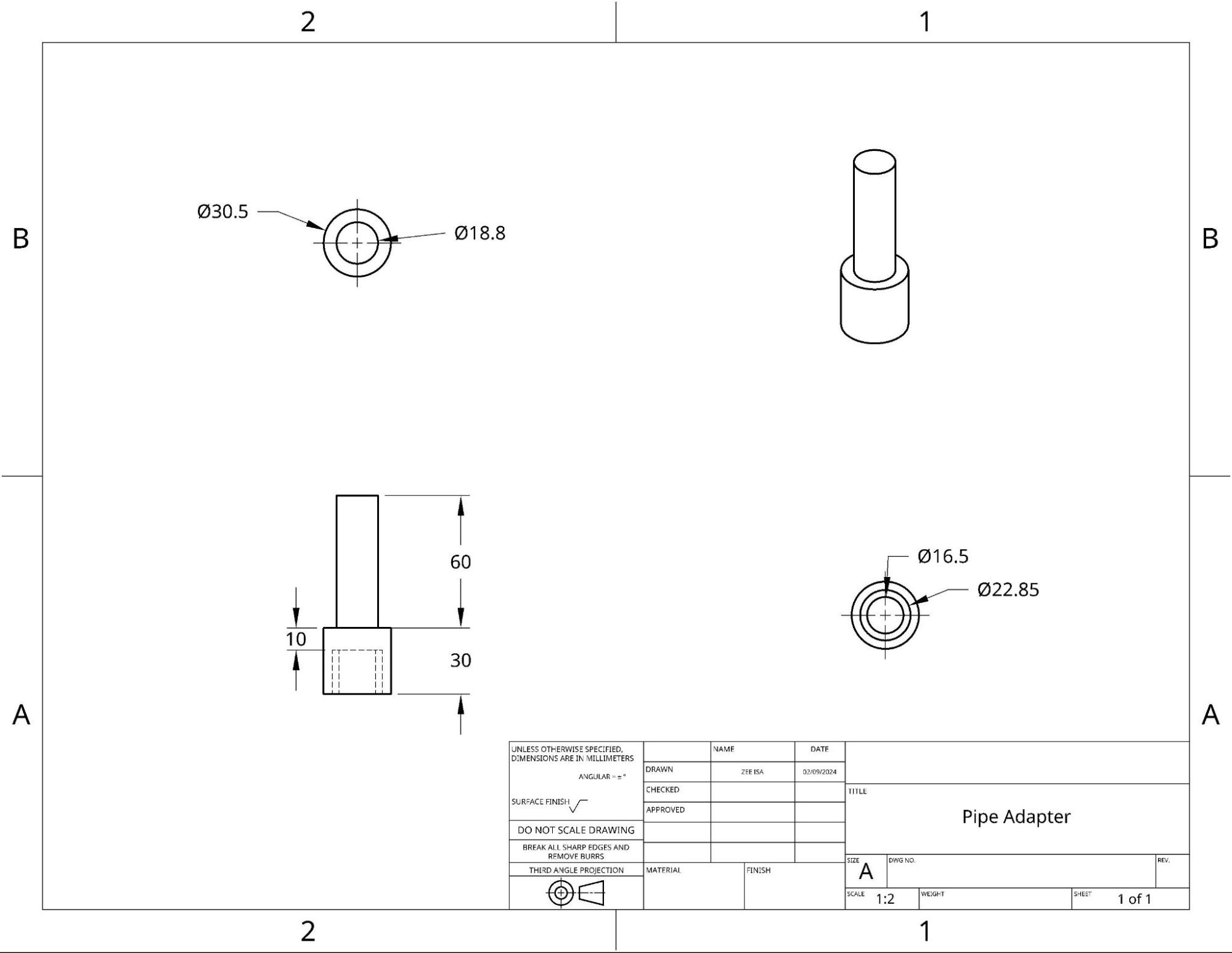}
\caption{Pipe adapter technical drawing}
\end{figure}

\begin{figure}[h!] 
\centering 
\includegraphics[width=8cm]{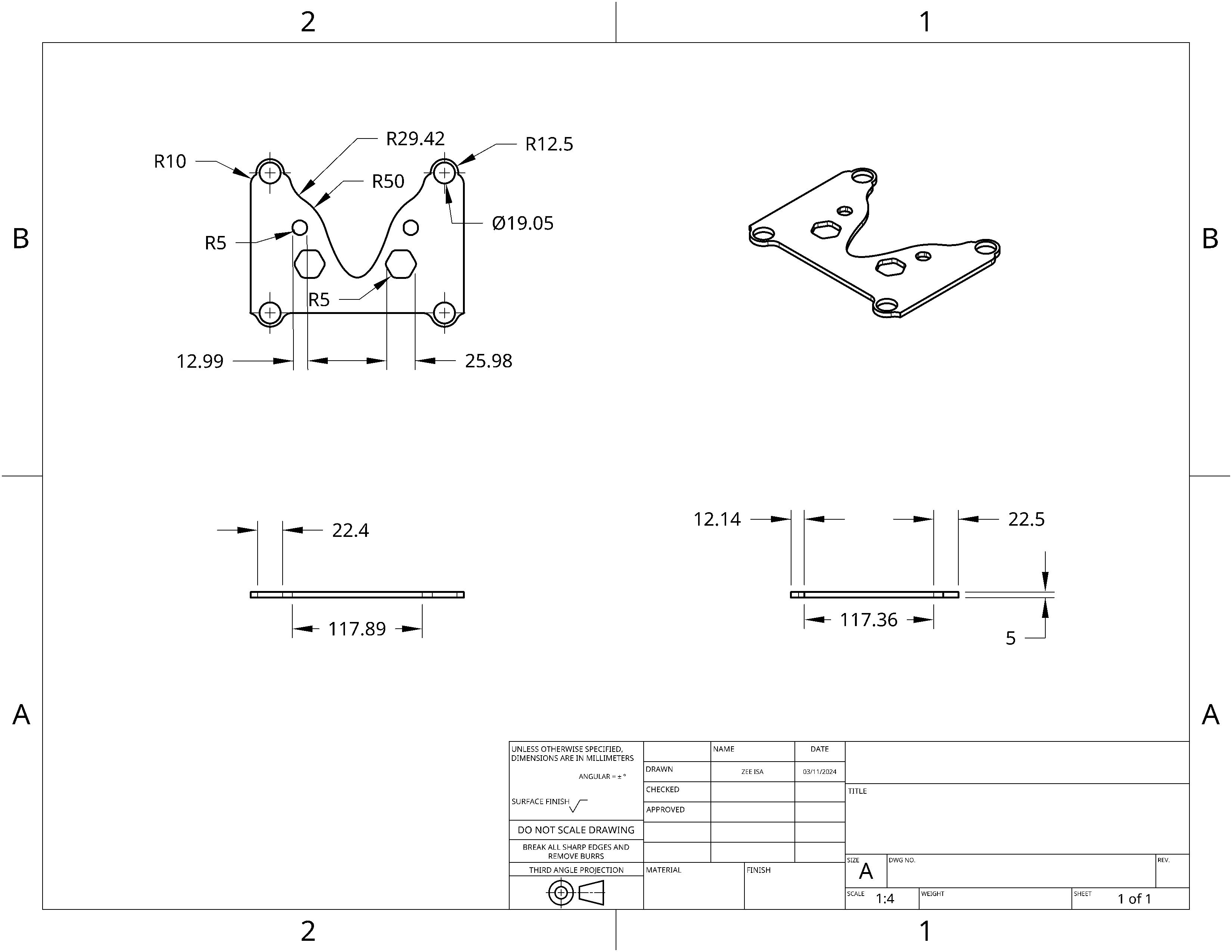}
\caption{Pipe aligner technical drawing}
\end{figure}

The final parts of the design are the clamps which hold the motors onto the trawl frame. These clamps have been modified from last year in two primary ways, both of which should greatly improve the clamps’ reliability and strength. The first change involves the mechanism by which the two clamps used for each motor are attached to each other. In the first iteration, they were bound together by zip-ties, which worked in the short term but were problematic at best for long-term use. This second version of the clamps uses 60mm bolts to hold together the clamps, allowing for greater clamp strength and significantly improved durability. In addition to this change, the clamps have been physically heightened to have a second point of contact with the trawl frame in order to prevent motor tilting and ensure that the thrusters remain perpendicular to the trawl frame even through collision. The technical drawings for each side of the clamp are shown on the next page. 

\begin{figure}[h!] 
\centering 
\includegraphics[width=8cm]{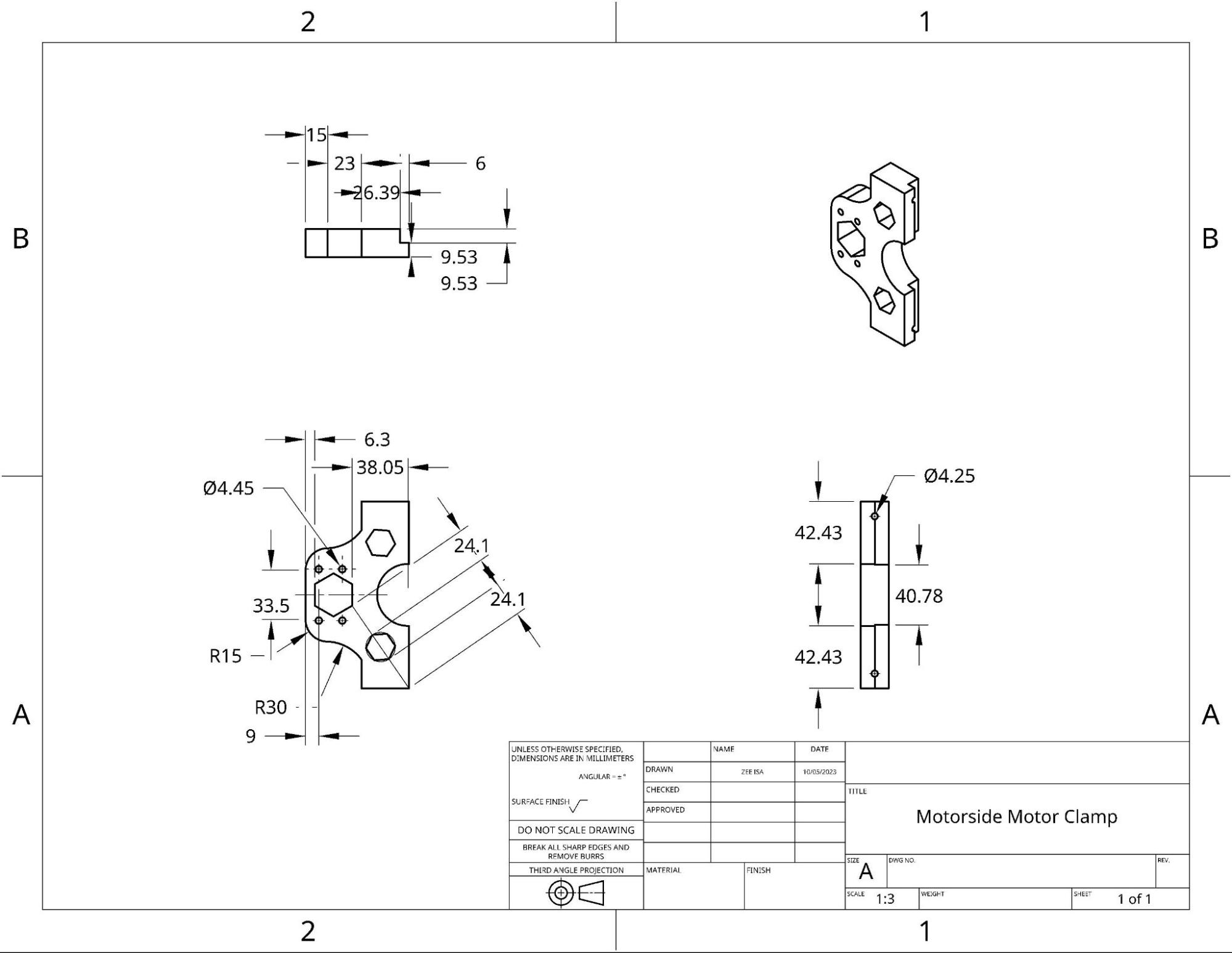}
\caption{Motor-side motor mount technical drawing}
\end{figure}

\begin{figure}[h!] 
\centering 
\includegraphics[width=8cm]{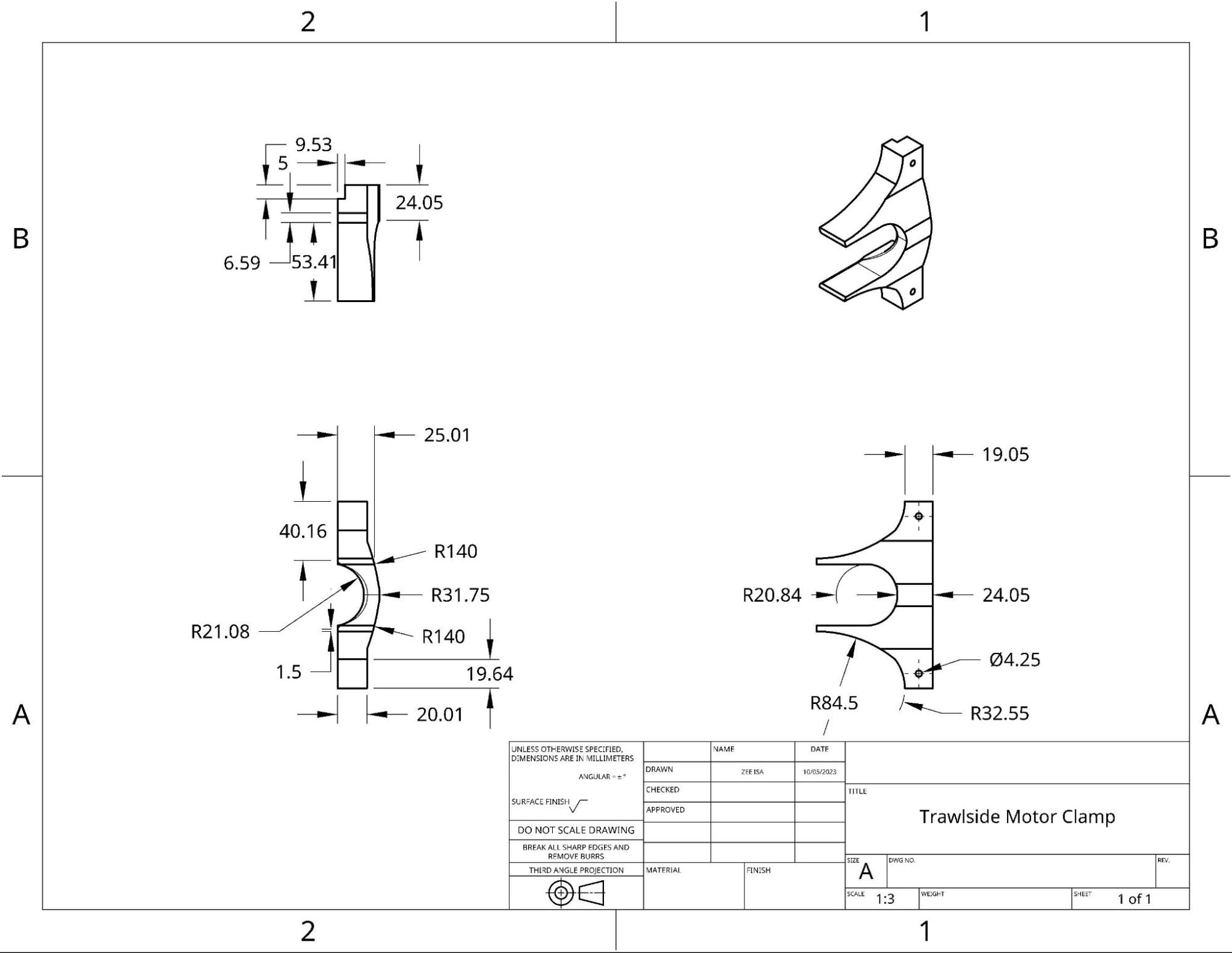}
\caption{Trawl-side motor mount technical drawing}
\end{figure}

The final aspect of the trawl’s construction is that of the buoys, motors, and net themselves. The net is attached using a nylon cord looped around the trawl’s frame and through the grommets placed regularly along the net. Simple square knots along with more nylon ropes are used to secure the buoys to the frame.

\subsubsection{Electrical Design}
As a part of the design process for this device, an electrical diagram was needed in order to properly document the creation process and facilitate simpler fixing of electrical issues. Below is the electrical schematic for the device as a whole. The schematic was created in KiCAD, an open source software for electrical design. 

\begin{figure}[h!]
\centering 
\includegraphics[width=8cm]{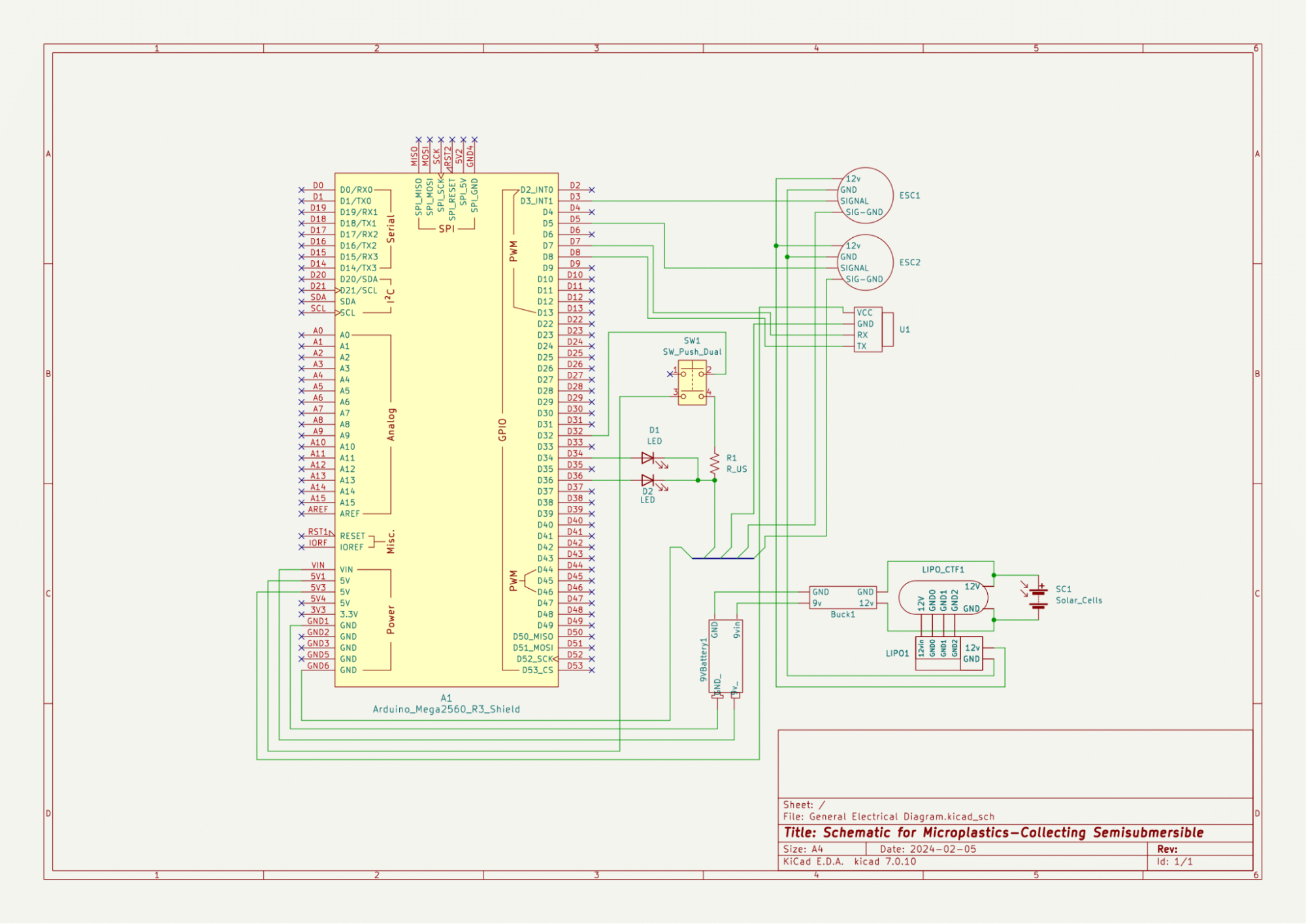}
\caption{Electrical schematic}
\end{figure}

In addition to creating an electrical schematic, empirical measurements of current draw under load were calculated from the device's testing in the Milwaukee River, and this along with empirical testing of the solar panel's power generation specifications allows for modelling of how the trawl's battery would fluctuate throughout a typical day of operation. Thus the below figure shows a graph of this, accounting for each of the aforementioned factors and assuming a charging period of 10h and using data collected from the National Solar Radiation Database.
\begin{figure}[h!]
\centering 
\includegraphics[width=8cm]{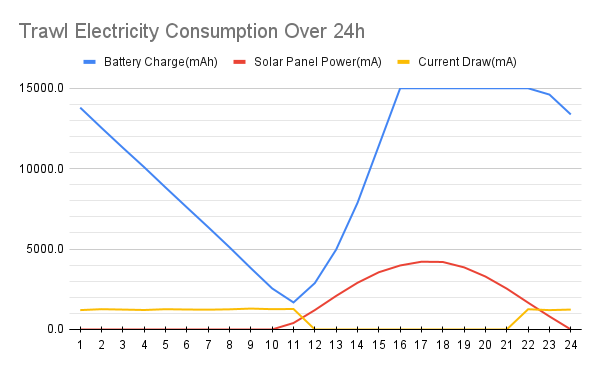}
\caption{Electrical schematic}
\end{figure}
\subsubsection{Stress Modelling}

The majority of pieces in this assembly experience negligible forces which are almost certain not to have a significant impact on the part’s structural integrity. The primary exceptions to this are the main control box and the side wings. To account for this, stress modelling was performed on these two pieces’ CAD models to predict the effects which various stresses would have on the pieces. This modelling was done within OnShape’s mechanical stress modelling application called CAEplex. The maximum stress experienced by the piece is 19,000 pascals, and the model bends very little from the stress exerted. The most significant stress is a normal stress in the z direction exerted by the central solar panel on the top of the box from the lid, and this stress is predicted to be less than 1\% of the maximum stress before breakage of the PLA. This is closer to the maximum stress of the piece than preferable, however, it is highly unlikely that the stresses experienced by the piece will double at any point during the trawl’s operation. The real-world predicted warping of the piece is shown below, along with warping exaggerated to be 250,000 times stronger than real-life to illustrate the primary areas where stress is being experienced by the trawl.

\begin{figure}[h!] 
\centering 
\includegraphics[width=8cm]{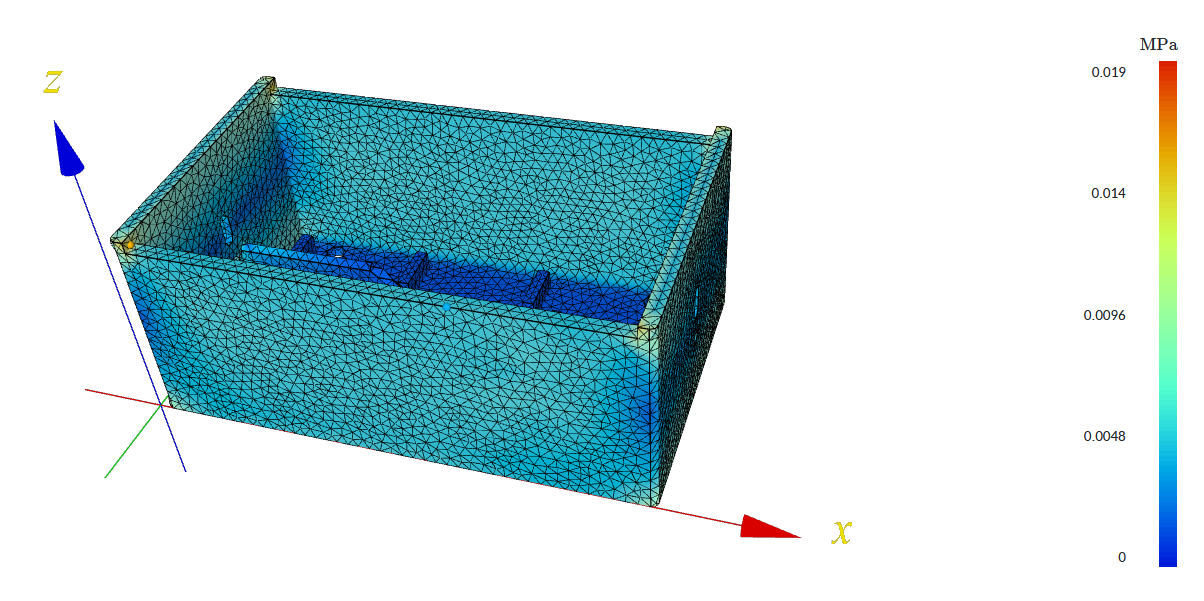}
\caption{Control box force analysis}
\end{figure}

\begin{figure}[h!] 
\centering
\includegraphics[width=8cm]{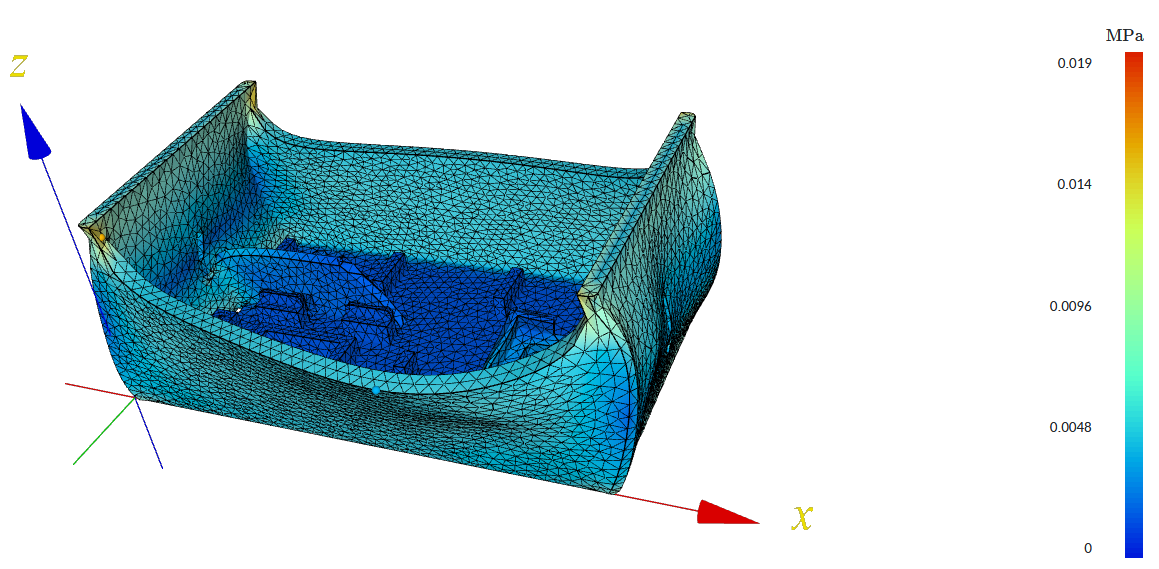}
\caption{Exaggerated control box force analysis}
\end{figure}

\subsubsection{Computational Fluid Dynamics(CFD)}
CFD analysis was performed utilizing SimScale along with Paraview for visualization. These results will be used to calculate the maximum forces exerted upon the trawl in order to determine the ballast mass necessary to prevent the trawl from flipping.
At first glance, conducting this type of analysis seems unnecessary, as it would be relatively trivial to simply add enough ballast to lower the trawl’s center of mass such that the force of the water will never exert any torque onto the trawl. Despite this, it must be considered that the entire control box assembly weighs nearly 10 kg by itself, and thus that adding enough weight to lower the trawl’s center of mass from just above the water’s surface to more roughly ¼ meter below it is impossible while still maintaining buoyancy and keeping the control box assembly out of the water. As simple proof of this, we can solve the center of mass equation using our known variables of the control box’s mass (roughly 10kg) its center of mass (roughly 10 cm above the water’s surface ideally), and the maximum depth at which ballast could be set (0.5 meters below the water). Thus: 

\begin{displaymath}
y_{cm}=\frac{m_{1}y_{1}+m_{2}y_{2}}{m_{1}+m_{2}}
\end{displaymath}

\begin{displaymath}
-0.25=\frac{0.1(10)-0.5w}{10+w}
\end{displaymath}

This resolves to:

\begin{displaymath}
w=14kg
\end{displaymath}

Given that the buoyancy of the trawl’s buoys is equal to roughly 17 kg of water displaced each, and that the total trawl in this scenario would weigh between 25-30 kg, the buoys would be just barely able to lift the trawl and the control box would be partially submerged for nearly the entire duration of operation, something which no amount of electronics waterproofing will protect against –at least at the scale and budget of this project. Therefore the most pertinent question to ask is not how much weight must be used to completely prevent the trawl from tipping, but how much it is acceptable for the trawl to tip. 
Conducting this analysis is important to the trawl’s success due to the challenges experienced in the first iteration of the trawl where flipping was very common even in relatively slow moving water due to a lack of ballast weight. Ideally, these CFD simulations can inform the decision as to what mass of ballast to use in order to minimize the weight gain of the trawl while maintaining acceptable tilt in ideal conditions. After conducting these simulations, several results were obtained. In the worst case, the trawl’s net is assumed to allow absolutely no water through it, causing a drag force equivalent to that of a flat paddle 1m wide and 0.5 meters tall. When conducting this analysis, many distinct states were examined, with the two most important being that of a vertical, flat drag approximation and that of a 45° approximation. The vertical simulation is the most intuitive of the two, with the trawl naturally resting roughly vertically when at rest, however, when in motion the trawl will tip to some degree due to the drag forces being simulated here, and thus simulating an arbitrary 45° tip gives an idea as to the range of forces which may be exerted upon the trawl during operation. Shown below are graphs showing the maximum moment of inertia when the worst-case representation of the trawl is traveling at 2m/s, just above its speed of propulsion.

\begin{figure}[h!] 
\centering 
\includegraphics[width=8cm]{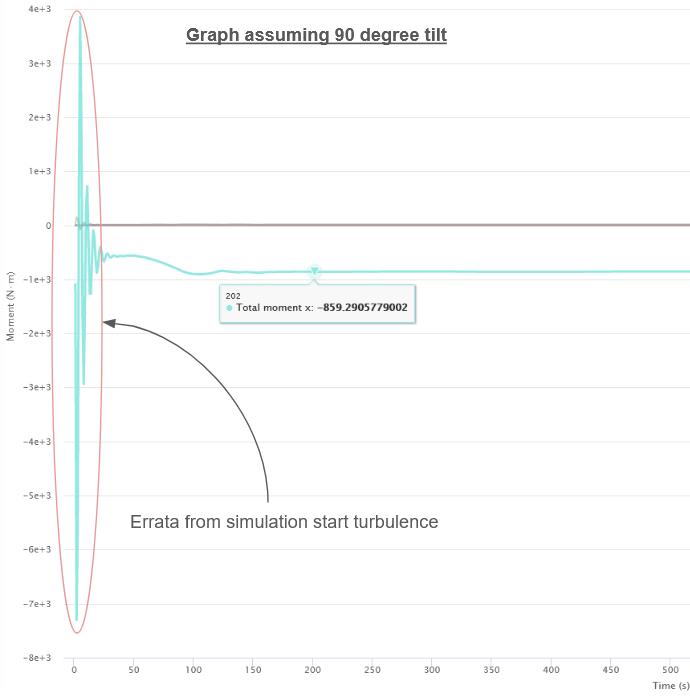}
\caption{90 degree CFD force graph}
\end{figure}

\begin{figure}[h!] 
\centering 
\includegraphics[width=8cm]{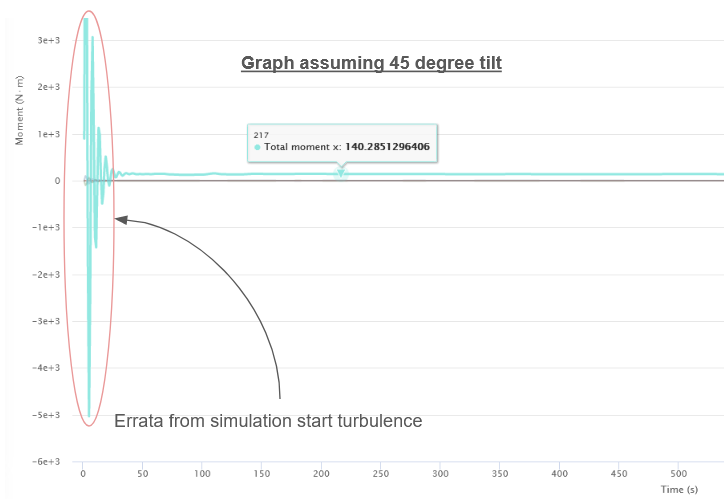}
\caption{45 degree CFD force graph}
\end{figure}

As can be seen in the graph immediately above, when at 90 degrees the moment of inertia is equal to roughly 860 Newton meters. To calculate torque, the equation \(\tau\) = r \(\times\) F can be used, giving a result of roughly 215 Newtons of force which must be counteracted in order to maintain stability. It is important to note that this force must not be constantly fought, as the greater the tilt imparted upon the trawl by it the lesser the force itself. This is demonstrated with the 45 degree approximation graph which shows that once the trawl has tilted 45 degrees, only 140 Newton meters of force are exerted upon it, and \(\tau\) = 35N. 


In order to counteract these forces, the simplest solution is to add weights to the bottom of the trawl. Using $F_{x}=mgsin(\theta)$, we find that an 80\(\degree\) equilibrium is reached at roughly 7 kg of weight at the bottom of the trawl. This is found using the following equations.
\[F_{weight}=9.8(m)sin\theta\]
\begin{multline}
    F_{water}=-967.5956+144.257x\\-7.648865x^2+0.1886765x^3-\\0.002154494x^4+0.000009248496x^5
\end{multline}

This equation for $F_{water}$ is obtained through a quintic regression of the six points obtained from a CFD analysis of the worst case scenario model of the trawl at various angles relative to the water’s surface. The x-coordinate of these six points represents this angle, and the y-coordinate represents the total force exerted on the trawl normal to its face.
\[\tau_{trawl}=\frac{1}{4}(F_{water})sin\theta\]
This torque equation accounts for both the force from the water and the angle which it is being exerted at. The $\frac{1}{4}$ comes from the fact that the center of mass of the trawl’s face is $\frac{1}{4}$ meters below the surface of the water in ideal conditions. 
All 3 of these equations can be seen in the following graph, where the x-axis represents the angle of the trawl relative to the water’s surface (90 is vertical) and the y-axis represents alternately $F_{weight}$ (blue line) and trawl (red line). The most important point on this graph is the point where the blue and red lines intersect, representing where the torque flipping the trawl is equal to the force exerted by the ballast at the bottom of the trawl when the trawl is at 90\(\degree\) relative to the water’s surface –the normal operating condition. The intersection of the yellow line $F_{water}$ and the blue line $F_{weight}$ is meaningless because it does not represent the actual force flipping the trawl, only the total amount of force on the trawl. 

\begin{figure}[h!] 
\centering 
\includegraphics[width=8cm]{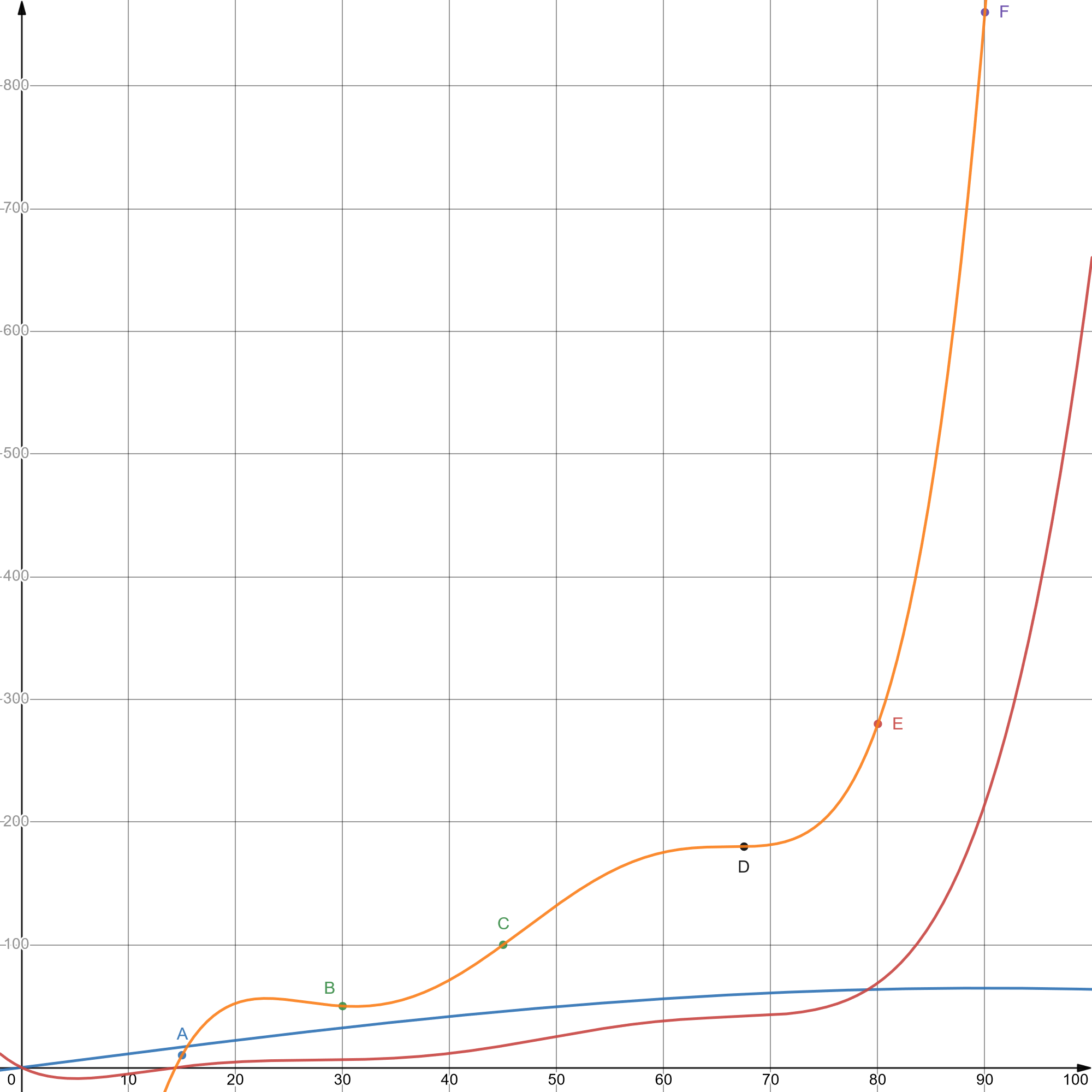}
\caption{Torque of water vs. anti-torque of ballast}
\end{figure}

In this case, as aforementioned, using roughly 7 kg as the value for the mass at the bottom of the trawl gives the device an 80\(\degree\) tip relative to the surface of the water, cutting the total weight to be added in half and greatly reducing the load on the buoys. Although this still allows for tipping in choppier waters, this tipping would reverse itself once waters became calmer due to the trawl’s weight below its two primary buoys (7 kg ballast + 5 kg frame) weighing more than the weight above the buoys. Below are shown visualizations of the CFD analysis performed.

\begin{figure}[h!] 
\centering 
\includegraphics[width=8cm]{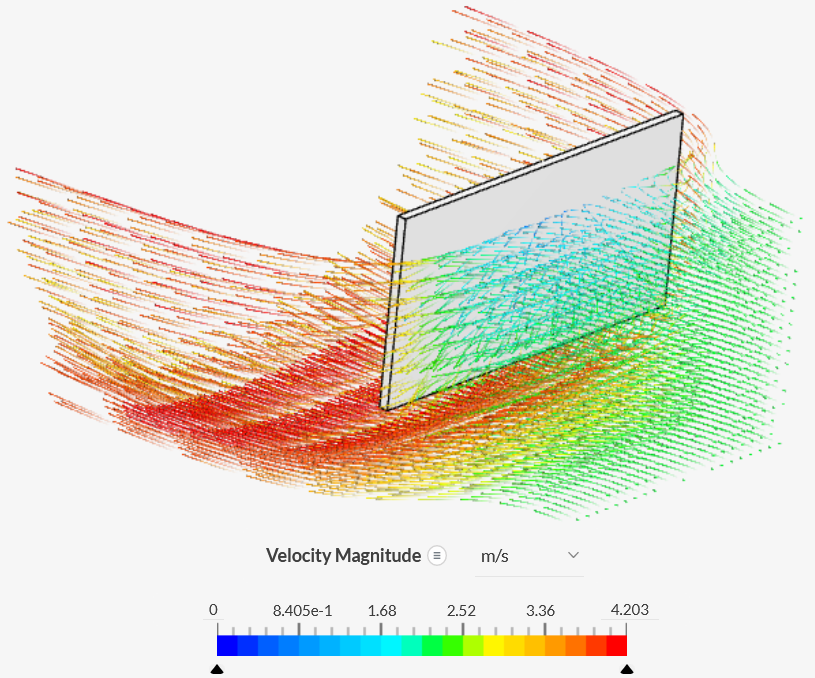}
\caption{90 degree CFD visualization}
\end{figure}

\begin{figure}[h!] 
\centering
\includegraphics[width=8cm]{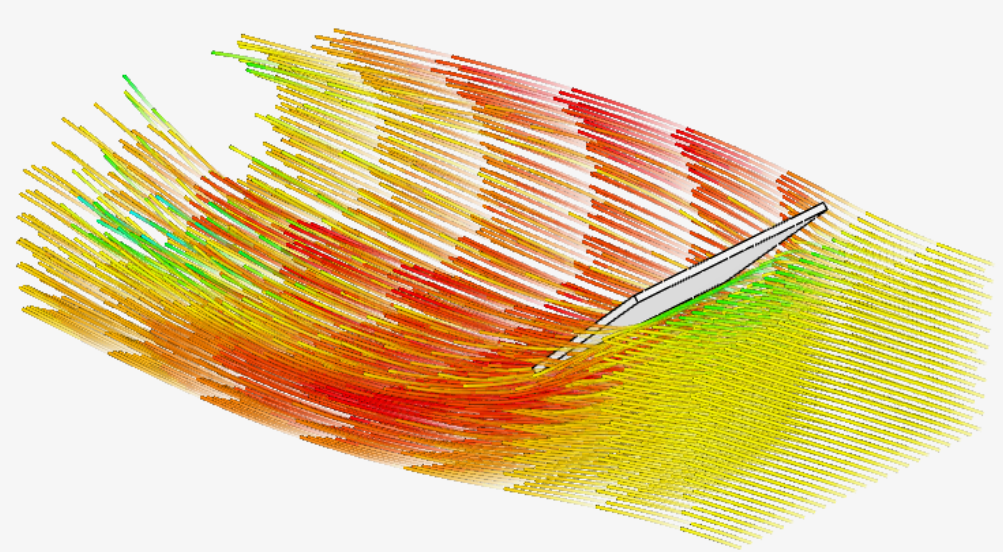}
\caption{45 degree CFD visualization}
\end{figure}

\subsubsection{Buoyancy and Weight Distribution}

As more and heavier components are added to the trawl, weight becomes a serious issue, especially with regard to the ballast necessary for stability in the second iteration of the device.  
A surprising and interesting consideration to make here is that of birds. Birds land on objects on the water relatively frequently, and whereas most boats and buoys may be unaffected by a bird’s weight, a much lighter device such as this may very well be. The largest bird active in the Great Lakes region is the American White Pelican, which can weigh up to 14 kilograms. Although this weight would not threaten the structural integrity of the device, it poses a significant threat to the buoyancy of the trawl, as the trawl’s buoys are limited in their lifting capacity. The frame of the trawl weighs roughly 3.2 kg, with the 3d printed parts adding roughly 4 kg altogether and the solar panel contributing roughly 3.5 kg. Along with the combined extra weight of the ballast and other miscellaneous components, this very generously totals to roughly 15 kg which is counterbalanced by a set of 4 spherical buoys which each have a circumference of 1 m, thus optimally displacing 16.98 kg of water each. This gives a margin of 52.92 kg, enough to comfortably handle most threats from birds, storms, or other dangers to the trawl’s floatation. Although this still allows the possibility of a very large bird landing destabilizing the trawl, the unlikeliness of such an occurrence as well as the waterproofing of the trawl’s electronics mean that any such destabilization would likely be temporary and low-risk. 
One other risk to trawl buoyancy is that of storms. The Great Lakes are large enough bodies of water that they form large waves and storms which have capsized many ships throughout history. To avoid this, the trawl will be made self-stabilizing by attaching the aforementioned ballast weights to the bottom of the trawl. This, along with the waterproofing of internal components of the trawl mean that the trawl should self-stabilize even after a major storm as long as structural integrity has been maintained. If integrity is lost, depending on which part of the trawl has been damaged, it is possible that the trawl could fail to stabilize or could sink. This is something of an unavoidable reality given the scale of the device, however, and not much can be done to definitively ensure that the trawl will retain complete integrity even through the harshest storms, given that multiple Great Lakes have recorded waves higher than 8 meters (20 feet) in the past. Despite this, the utmost effort has been made to ensure that the device is structurally sound within the constraints of the weight, resource, and monetary restrictions of the project.

\subsection{Wave Flume Machine Testing}
The final aspect of testing the device comes through testing in a Wave-Flume Machine (WVFM). WVFMs are devices capable of simulating very many different wave conditions on a smaller and far more controlled scale than that of a real lake or river, permitting significant experimentation without danger while still under realistic conditions. The trawl in this project was tested in the WVFM at Michigan Technological University in Houghton Michigan called MTU Wave, a 10 m long, 3 m wide, and 1 m deep WVFM capable of generating waves of up to 30 cm in amplitude (60cm crest-trough). Here, 6 rounds of testing were conducted using waves of varying amplitude, frequency, and regularity ranging in amplitude from 3 cm to 7 cm for regular wave patterns along with waves of amplitudes up to 20 cm when testing the irregular JONSWAP wave spectra. The WVFM was not run at full amplitude due to other research projects which were susceptible to damage if impacted by such waves. The JONSWAP spectra is a waveform which closely models the real-life behavior of waves in large bodies of water, thus it will be the test most closely analyzed here. Data collection was conducted in a few ways, with the primary data being video captured by a camera stationed at one end of the WVFM along with corresponding data collected from wave height gauges located roughly 1 m in front of the device during testing (only rough measurement is possible because the trawl moved significantly during testing). These data points can be corresponded to examine the impacts which various wave amplitudes and regularities have on the trawl’s stability. 
The wave amplitudes examined here are typical of days without storm on the Great Lakes, with typical significant wave height (the average of the top 33\% of wave heights) lying around 20-30 cm in amplitude as shown by the below graph of historical wave data accessed from NOAA data buoy \#45007. 

\begin{figure}[h!] 
\centering 
\includegraphics[width=8cm]{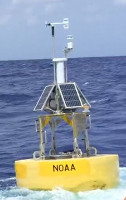}
\caption{NOAA buoy \# 45007}
\end{figure}

\begin{figure}[h!] 
\centering 
\includegraphics[width=8cm]{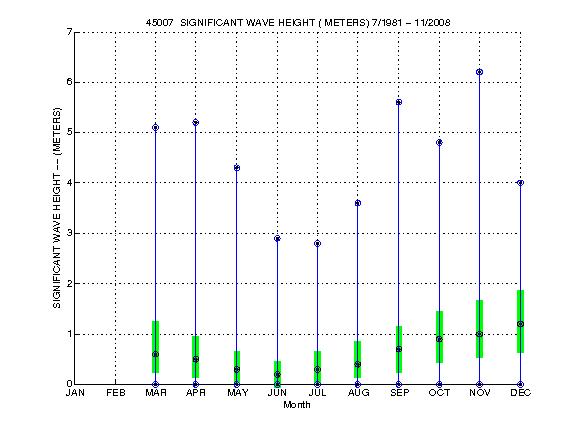}
\caption{Significant wave height over time}
\end{figure}

As can be seen from this box-and-whisker of significant wave height in meters, the average of the median significant wave heights average out at roughly 0.5 m (amplitude of 0.25 m) across the entire year, meaning that conditions in the JONSWAP test will quite closely mirror those seen in the actual testing environment. Conditions will exceed this estimate for roughly half of the time in which the trawl operates, however, the trawl should be able to collect microplastics with similar effectiveness even in inclement weather as long as the buoyancy and integrity of the trawl are maintained. Below is shown the graph of the wave amplitudes over time during the JONSWAP testing. 

\begin{figure}[h!] 
\centering 
\includegraphics[width=8cm]{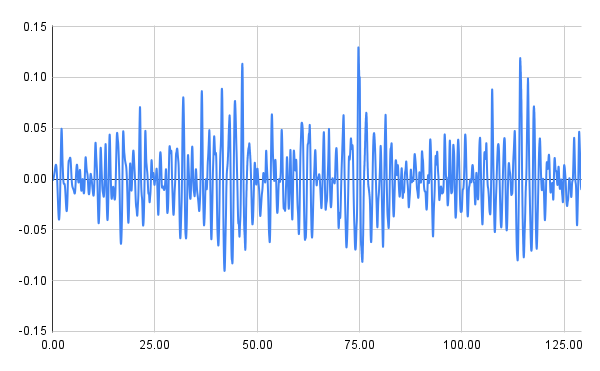}
\caption{JONSWAP wave amplitudes}
\end{figure}

\subsection{Risks}
This project is relatively low-risk physically, excepting the standard risks of working in a workshop environment. The primary tools used will be additive manufacturing processes such as 3D printing and simple constructive tools like screwdrivers, wrenches, pliers, and scissors. In addition there will be some very minor usage of basic power tools such as drills. These tools and activities present very little risk to an experienced operator, and they will be operated primarily in a school under the supervision of an engineering teacher.
Environmentally, the trawl uses a prefilter of wider pore diameter to ensure that macro-organisms such as fish do not enter the trawl, however, given the similarity of size, shape, and density of microorganisms and microplastics there is no guarantee that some microorganisms will not be uptaken by the net. This is predicted to be of very minimal impact on the small scale of this project’s testing, however a more full analysis of potential impacts to the environment and the consultation of professionals in the field will be necessary if the project is to be deployed on a larger scale as described in the following sections.

\subsection{Cost Analysis}
Deploying these trawls would be a significant endeavor in and of itself, however the process could be aided by following microplastic distribution patterns and lake current maps in order to collect plastics most efficiently. The cost per unit is roughly \$1,000, with the metal tubing costing \$458, the mesh costing \$285, Arduino Mega \$27, the two motors \$64 each, the solar panels \$22 each, 3D printing material roughly \$25, the battery \$140, the buoys \$36 (\$18 each), the sonar \$14, PVC pipe roughly \$7, and other assorted costs such as wiring, PVC pipe for the cod-end of the trawl (small section at the end of the trawl to hold collected plastics), and other miscellaneous expenses amounting to roughly \$25. In summation, the product created has a total cost of \$1,115 (with a fully self-sufficient version including solar panels costing roughly \$1,115) and an efficiency (upon observation) equal to the efficiency of boat-hauled Manta Trawls based upon rate of travel. Thus with an investment of roughly \$810,000 in material costs along with costs of deployment, i.e. using boats to place trawls into open water, labor, etc, this solution could be implemented, a number much smaller than one might expect for the cleansing of an entire Great Lake (Lake Erie used for this example). 

\section{Deployment}
Deploying the modified Manta Trawls used in this project into large bodies of water would likely be quite a challenge, primarily in ensuring that each trawl was placed precisely in order to collect the maximum number of microplastics possible. This would be most easily accomplished by deploying in a body of water with well-mapped or predictable currents, with the most easily notable example being the Great Lakes. 

\begin{figure}[h!] 
\centering 
\includegraphics[width=8cm]{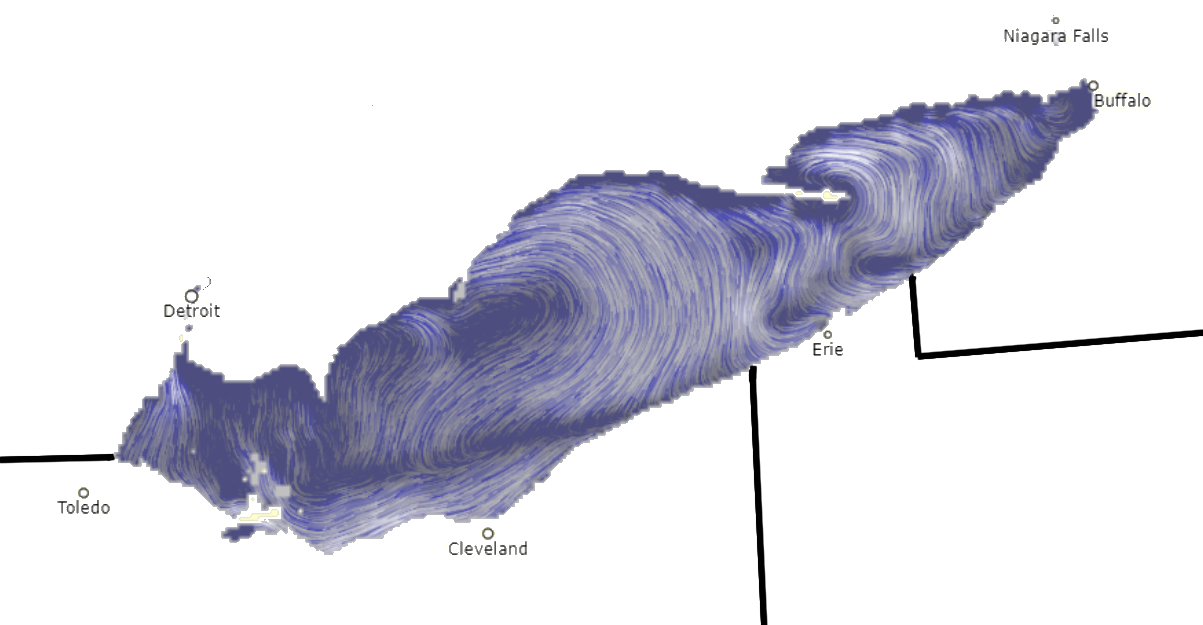}
\caption{Surface flow in lake Erie}
\end{figure}

\begin{figure}[h!] 
\centering 
\includegraphics[width=8cm]{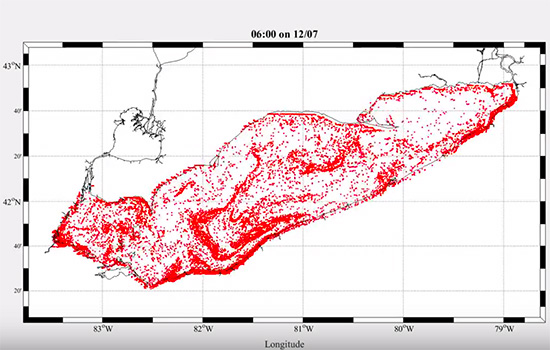}
\caption{Microplastic distribution in lake Erie}
\end{figure}

Above is an image of the currents throughout lake Erie, and in this particular case it is clear that nearly all of the lake’s currents converge into a single line, which one would expect to be nearly flooded with microplastics. Indeed, when viewing a map of the microplastic distribution within Lake Erie, one finds this to be true. The above model from the Rochester Institute of Technology shows that, as expected, the distribution of microplastics throughout Lake Erie follows the currents nearly exactly, and when the two maps are overlaid we see a clear image. 

\begin{figure}[h!] 
\centering 
\includegraphics[width=8cm]{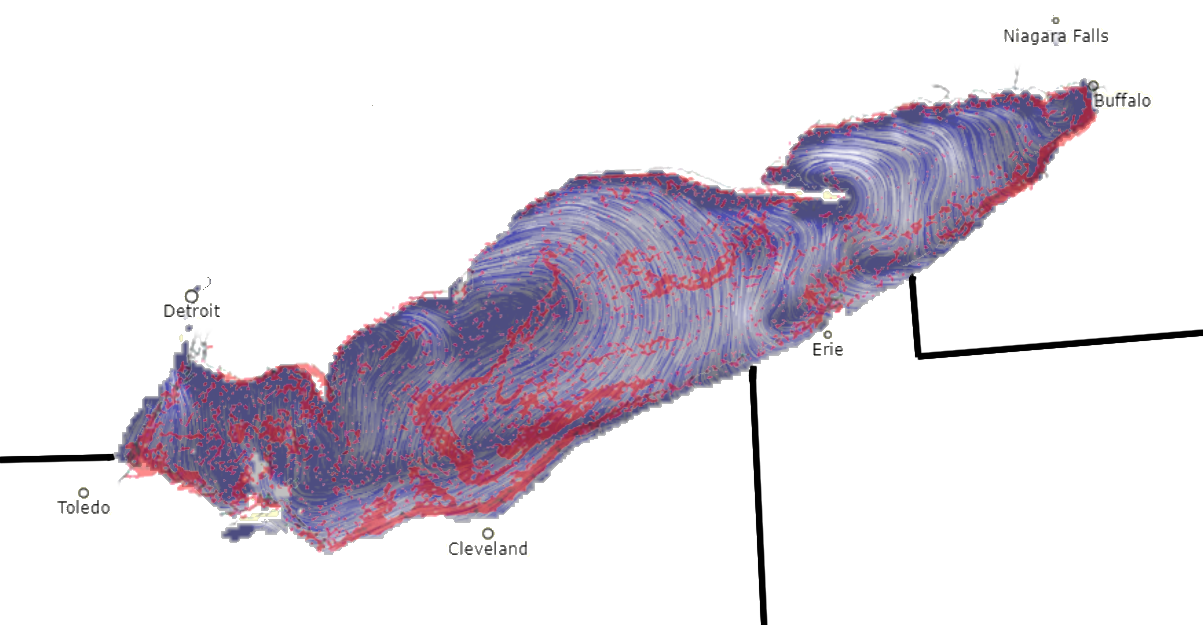}
\caption{Overlaid lake Erie flow and distribution maps}
\end{figure}

The two maps overlay nearly perfectly, and as one would expect, nearly all of the lake’s plastics are concentrated just north-northeast of Cleveland, where the currents of the lake converge. The two exceptions to this are in the area southeast of Detroit, where slow moving water traps microplastics entering from Lake Huron, and in the area southwest of Buffalo, where currents once again converge on the lakes southern shore in that area. This distribution of currents gives clarity to where the modified Manta Trawls should be deployed, and if used in these areas a single Manta Trawl could have a far greater impact than it could if used in less polluted areas. This example uses Lake Erie, however each of the Great Lakes have similar current maps, and when put together they create this visual.

\begin{figure}[h!] 
\centering
\includegraphics[width=8cm]{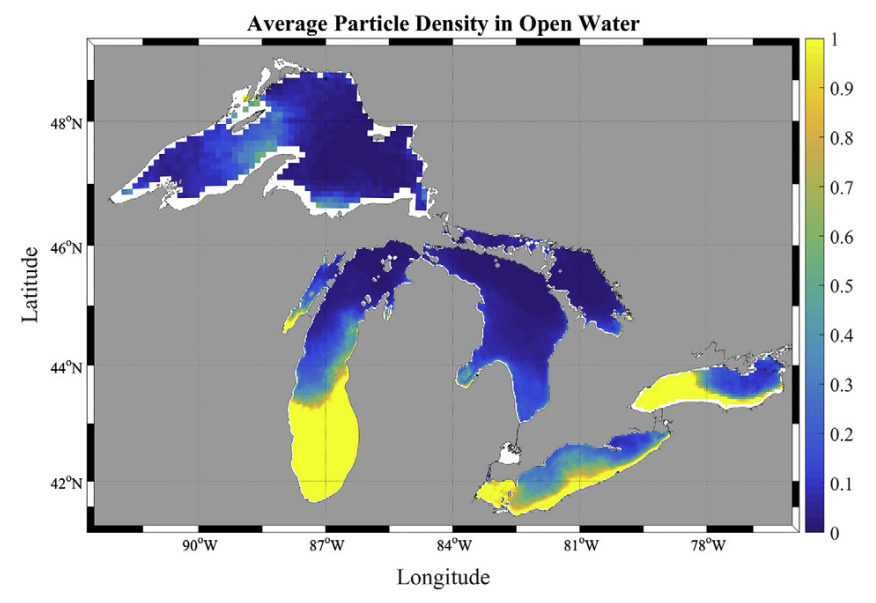}
\caption{Great Lakes microplastic particle distribution}
\end{figure}

This visual makes clear on a larger scale that microplastics in the Great Lakes (and likely in nearly all large bodies of water with consistent current patterns) are concentrated in an area only comprising roughly 22\% of the total area within the lakes. Given this information, along with the fact that this 22\% of the Great Lakes’ area is concentrated nearly entirely within the lower $\frac{1}{3}$ of the lakes Michigan, Erie, and Ontario, makes the issue of microplastics much more surmountable, and lends credence to the idea of ridding the Great Lakes of surficial microplastics. In summation, the currents of the Great Lakes and other large bodies of water can be significantly leveraged in order to most effectively deploy plastic-collecting Trawls, and collecting plastics from the Great Lakes is very feasible due, as well, to these currents.

\section{Initial Modeling}

\begin{figure}[h!]
\centering 
\includegraphics[width=8cm]{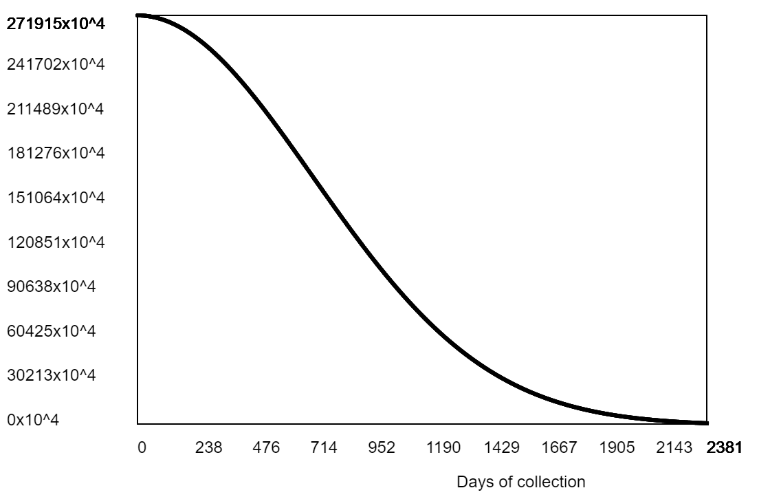}
\caption{1-Day deployment model: Erie}
\end{figure}

The above graph shows a model of the quantity of microplastics within Lake Erie given current predictions of microplastic concentrations and the assumptions that the modified Trawls created in this project are deployed at a rate of one per day into the lake. This model’s x and y scales represent the total estimated quantity of plastic particles and the number of days for which modified Manta Trawls are deployed at a rate of one per day and used constantly. The graph accounts for lowering concentrations of plastic as more and more plastics are taken from the lake as well as the growth in the number of plastics from direct pollution and from entry through Lake Huron. The model makes several significant assumptions, most effectually that microplastics are evenly distributed throughout the lake, which is known to be false, and that each of the deployed trawls operates for 12 hours each day, without breaking, leaching plastics to their surroundings, or limiting effectiveness in any other way. These are significant assumptions, and thus it must be acknowledged that the model presented is likely a very optimistic one. Despite this, as demonstrated in the previous section of this paper, most of the microplastics in Lake Erie (and the Great Lakes in general) are concentrated in a quite small, and with the exception of central Lake Michigan quite littoral (near-to-shore) area. Thus there are some significant factors that this model cannot predict, and thus the model should be taken primarily as a simple demonstration that lowering the concentration of plastics within large bodies of water is feasible, not necessarily as an indicator of exactly how long one could expect a lake to be near-completely cleared of plastics. The following model uses each of the same parameters, makes each of the same assumptions, and accounts for all of the same things, however it assumes that trawls are deployed at a rate of one per week as opposed to one per day. The model above ends the graph with 2,381 trawls deployed in the lake, and the model below ends with 802 trawls deployed in the lake, thus showing that if one is willing to wait longer, a far smaller number of trawls could be used to reach the same outcome as a much larger quantity. Below the one-week deployment model, there are labeled models for microplastic collection in Lake Michigan and Ontario (based upon the one-week model). Lake Huron is not modeled due to a very short water cycle and very low concentration of microplastics, Lake Superior is also not modeled due to incredibly low concentrations of microplastics due to a lack of major settlements on Superior’s shores. 

\begin{figure}[h!] 
\centering 
\includegraphics[width=8cm]{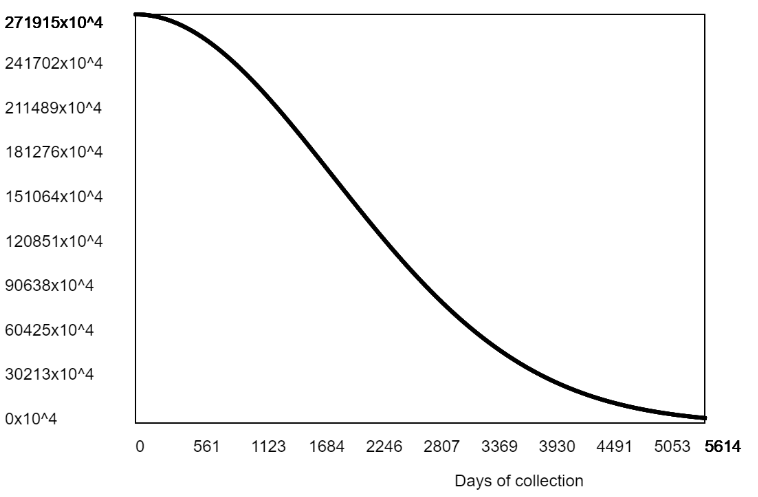}
\caption{7-Day deployment model: Erie}
\end{figure}

The graph below shows a model of micro plastic quantity in Lake Michigan, taking into account all of the variables which were mentioned in the models of Lake Erie.            

\begin{figure}[h!]
\centering
\includegraphics[width=8cm]{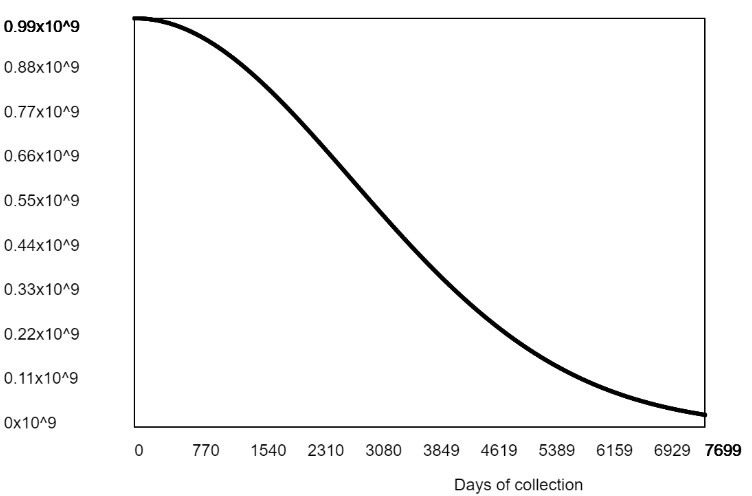}
\caption{7-Day deployment model: Michigan}
\end{figure}

\begin{figure}[h!] 
\centering
\includegraphics[width=8cm]{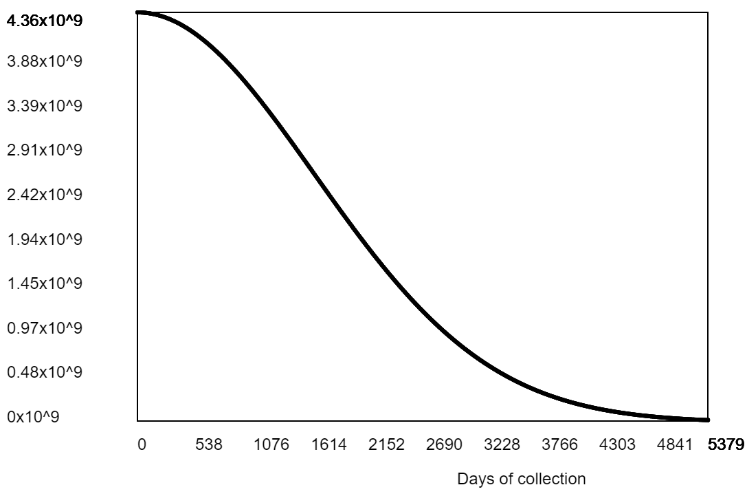}
\caption{7-Day deployment model: Ontario}
\end{figure}

This graph immediately above, similar to the other graph above, shows a model of microplastic quantities in Lake Ontario with the assumption of deploying trawls at a rate of once per week. Both of these graphs are displayed here simply as a demonstration of the effectiveness of this solution across several different scenarios. These results are, once again, very reassuring confirmations that this solution works in a variety of different circumstances, especially considering the fact the these models assume completely random distribution of trawls, which would be very inefficient, given that the microplastics in the Great Lakes are generally concentrated heavily within a single half or third of each lake. 

\section{Expectations of Trawl Effectiveness}

After developing the models shown in the \emph{Initial Modelling} section, the device was tested empirically in the Milwaukee River on the 20th of August, 2023. This testing was performed in the same location as one of the studies examined in this project, \emph{Vertical Distribution of Microplastics in the Water Column and Surficial Sediment from the Milwaukee River Basin to Lake Michigan}, at Milwaukee’s Estabrook Park (referred to as MEP in the study). This study was one of the primary sources used in creating the models in the previous section due to its consistent documentation of microplastic concentrations and locations as well as the ease of access to the study’s focus locations due to proximity and public access. In addition to this, the location of the study was also the location of a USGS gage-height sensor, a type of sensor measuring the height of a body of water (in this case the Milwaukee River) above a fixed point just below the riverbed. This data on water levels and total river discharge can be accessed online on the USGS website and can be utilized along with satellite imagery to create an estimation as to the quantity of water which passed through the trawl during its time on the river (\emph{Milwaukee River at Milwaukee, WI, 2023}). This was used in conjunction with data on the cross-sectional area of the river at the testing location. This data can then be combined with microplastic concentrations calculated by the aforementioned study on the vertical distribution of microplastics to create an estimate as to how many microplastics should have been collected by the device during the hour of its operation. The math to calculate this value is as follows:

\begin{enumerate}
    \item Calculate the amount of water passing through the manta trawl per second

\begin{enumerate}
        \item Find total river discharge at time of experiment - 423 ft\textsuperscript{3}/s (\textit{Milwaukee River at Milwaukee, WI}, 2023)
        \item Find approximate depth of river using depth sensor - 1.2 ft (\textit{Milwaukee River at Milwaukee, WI}, 2023)
        \item Use satellite imagery of location with depth to find rough cross-sectional area of river in the area of testing - 1.2 ft average depth * 106 ft wide = \(\approx\)127 ft\textsuperscript{2} cross-sectional area
        \item Divide total discharge (423 ft\textsuperscript{3}/s) by cross sectional area - 423 ft\textsuperscript{3}s / 106 ft\textsuperscript{2} = 3.99 ft\textsuperscript{3}/ft\textsuperscript{2}/s
This unit measures the volume of water passing through a 1’x1’ area perpendicular to the shore in one second. 
        \item Multiply this by the total area of the trawl’s opening - 5.38 ft\textsuperscript{2} * 3.99 ft\textsuperscript{3}/ft\textsuperscript{2}/s = 21.47 ft\textsuperscript{3}/s
        \item At this point we convert this value to the metric system for the sake of simplicity \\15.82 ft\textsuperscript{3}/s $\rightarrow$ 0.61 m\textsuperscript{3}/s
        \item Multiply this by 60 twice to convert from m\textsuperscript{3}/second to m\textsuperscript{3}/hour - 2,196 m\textsuperscript{3} of water passing through the net during one hour in total. This number seems very large, however we can make a quick check of the numbers by calculating the water speed from the known variables as follows: 
3.99 ft\textsuperscript{3}/ft\textsuperscript{2}/s means that the water is traveling at \(\approx\)4 ft/s. This converts to a range of 2-3 mph, quite a reasonable speed for a slow moving river.

        \item As per a previous study in this same location, the microplastic concentration in the river should be roughly 1.58 p m\textsuperscript{-3} (0.48 particles per cubic meter).
        \item Multiply this by the total volume of water which was filtered by the trawl - 2196*1.58=\textbf{3469 }
Thus, we would expect this device to collect something in the range of 3,123 - 3,815 microplastics in one hour of collection, assuming the exact same conditions as the test performed in the previous study \emph{Vertical Distribution of Microplastics in the Water Column and Surficial Sediment from the Milwaukee River Basin to Lake Michigan}. As a note, this range is +/- 10\% of the original expected number. 10\% is arbitrary due to a lack of standards.
\end{enumerate}
\end{enumerate}

 This number therefore seems quite reasonable when compared to other studies which have also collected similar numbers of plastics in microplastic-polluted areas \cite{Uurasjarviconcentrations}. Some studies have even found concentrations of microplastics over 30 p m\textsuperscript{-3} in waterways near populated areas \cite{BordosCarpathian}. The most important part of this estimated number of microplastics, however, is how it compares to the number actually collected during the roughly one hour long operation of the trawl created for this project. This will be examined in the next section. 

 \section{Results of Empirical Testing in the Milwaukee River}
 As calculated in the previous section, the trawl was expected to collect somewhere between 3,300 - 3,700 microplastics during its one hour of operation in the Milwaukee River. Exceeding this expectation, the trawl collected a total of 4,438 microplastics in roughly one hour of operation. This is roughly 127\% of the expected yield, a somewhat more significant increase than what one would expect, however a number of factors explain this incongruence quite well. The most important factor in this is likely the time which has passed since the study which the expected number was based upon. That study was conducted in 2019, roughly 4 years before this test was completed. This means that it is incredibly likely that the concentration of microplastics has changed to some degree in the time since. Increases of population, new developments, and changes to river hydrology all mean that the concentration of microplastics has almost certainly increased since the study was conducted in 2019. In addition to this, the 2019 study was conducted by wading through the water and holding the trawl in a single, fixed position whereas this test was conducted using a moving trawl. This movement would have increased the volume of water passing through the trawl which would have correspondingly increased the total number of microplastics observed. Finally, lacking the materials and equipment necessary to completely separate the microplastics from organic matter such as algae and other microorganisms in the river meant that when counting the plastics there were almost certainly several misidentifications which likely led to counting organic material as microplastics and vice versa. Along with these factors which could have increased the concentration of microplastics in the water, the test was also conducted with the trawl’s entire mouth below the water’s surface as opposed to being only halfway under in the aforementioned study. This likely caused the sampling of many neutrally or slightly negatively buoyant plastics as evidenced by the images of the samples in the next section. According to that same study, samples taken at the surface at the site of the testing slightly underestimate the average concentration of plastics throughout the entire water column as seen in the graph below retrieved from that study. 
 
\begin{figure}[h!] 
\centering 
\includegraphics[width=8cm]{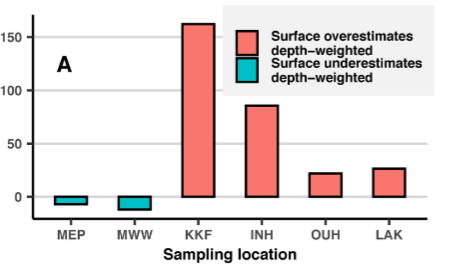}
 \caption{Depth weights on microplastic sampling}
\end{figure}

 The sampling location for both the study and this project’s test were within several hundred feet of each other, and the sampling location in the study is referred to as MEP, the first bar in the graph. This means that the slightly deeper sampling of this test would have collected a slightly higher concentration of plastics than the shallower tests done in the previous study.

 Overall, however, with the experience of counting the plastics, it seems that these errors would have been quite rare due to the utilization of several people to independently count microplastics from the same samples in order to ensure that one individual did not significantly over or undercount the microplastics. Prior studies have found that visual analysis like this has an error rate of roughly 20\%, meaning that the initial figure of 4,438 microplastics likely ranges from 3,550 to 5,326 microplastics, with the lower end of that range being very close to the predicted number of 3,469 microplastics.

 \section{Classification and Counting}

 After collection of the plastics in the river, the trawl was held vertically (mouth up) and sprayed with tap water in order to wash all plastics into the cod-end of the trawl (detachable section at the end of the trawl).  After this, the cod-end was removed and further washed with tap water into several clear glass containers for counting. In lieu of the chemicals needed to separate the organic particles in the samples or the equipment needed to conclusively determine the identity of different particles, the particles in this test were counted manually through separation into various clear glass containers which were subsequently gridded into 1in x 1in squares and counted by square. This gridding allowed the plastics to be counted in a much more manageable way, with few squares containing more than 100 plastic particles, increasing counting accuracy and removing the possibility of forgetting where one was counting, especially important as the counting process took several hours. 
 
\begin{figure}[h!] 
\centering 
\includegraphics[width=8cm]{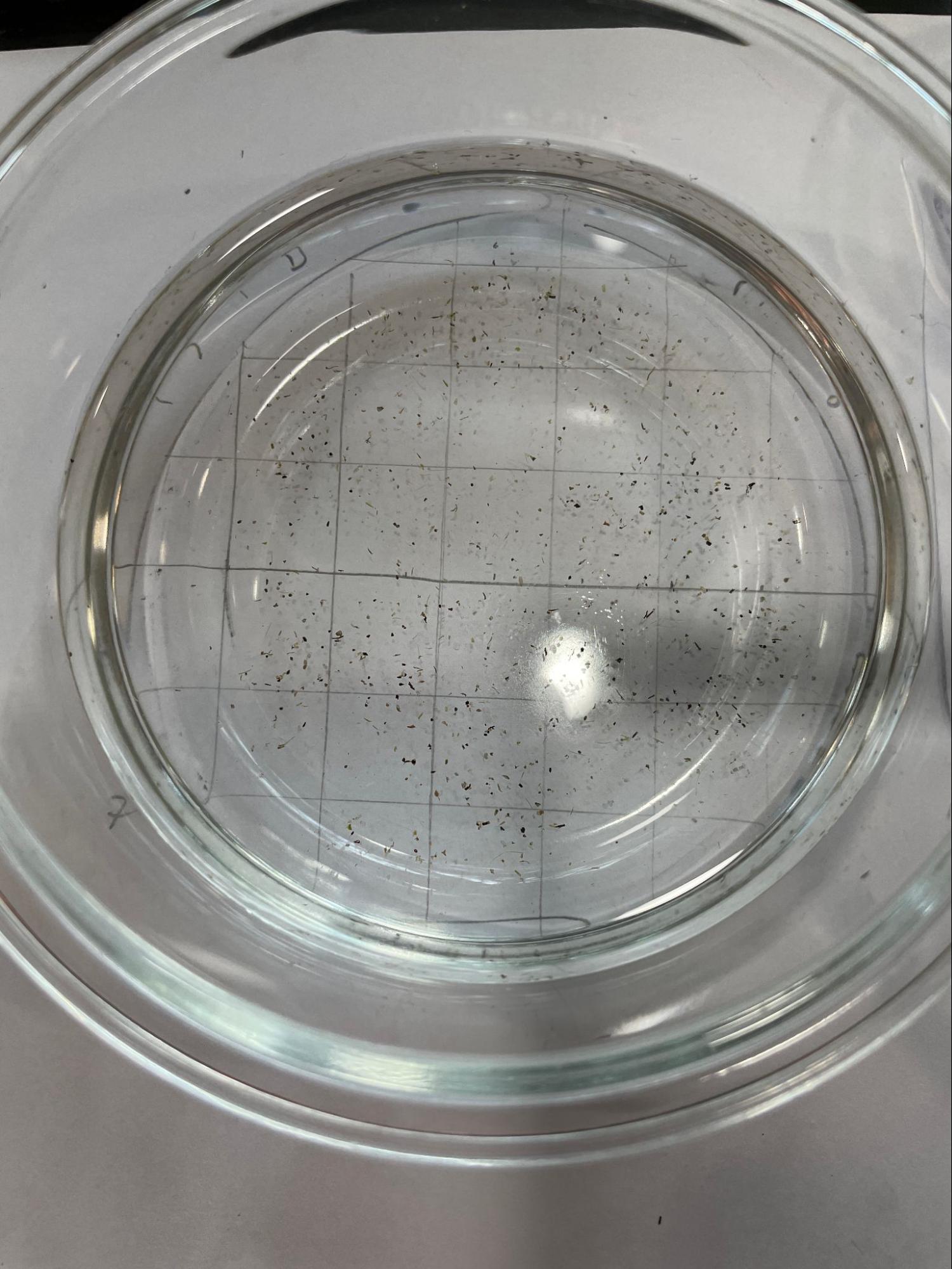}
 \caption{Sparse microplastic sample bowl}
\end{figure}

 Above is an example of a sample container with the grid below it. After separating the plastics into these containers, a thin rod was used to count the plastic particles, only counting particles which met the criteria set out by the study \emph{Analytical methods for microplastics in the environment}: 1) no visible cellular or organic structures, 2) unsegmented, 3) fibers of homogenous width (not tapered) and at least two of the additional criteria: 1) unnaturally coloured or with a brightly coloured coating (e.g. bright orange, blue etc.), 2) appear to be of homogenous texture/material, 3) abnormal (un-natural) shape e.g. perfectly spherical, 4) fiber that remained unbroken if tugged with tweezers, 4) reflective/glassy, 5) flexible without being brittle. The results for each square were then entered into a spreadsheet with each square colored according to the concentration of plastics within it. Before counting, the plastics were separated by buoyancy in order to transform the counting from a 3D search into a 2D one, further mitigating the chances of error. The spreadsheet for the container above is shown below.

 \begin{figure}[h!] \centering \includegraphics[width=8cm]{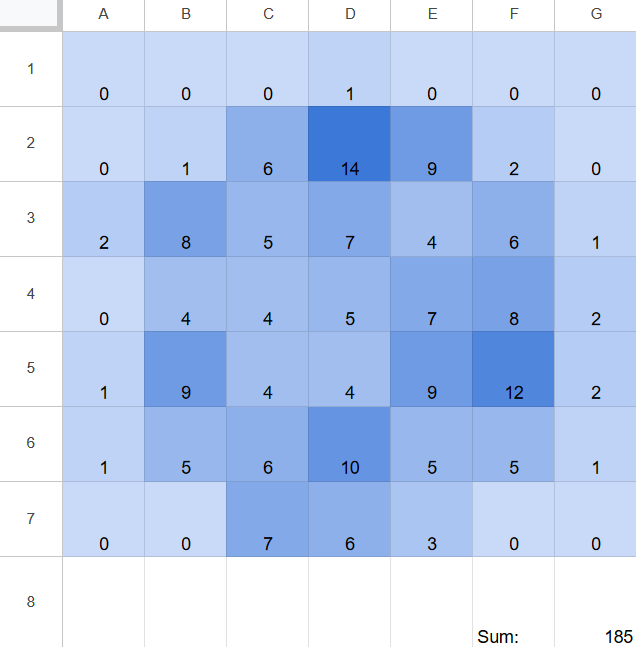}
 \caption{Sparse microplastic sample spreadsheet}
\end{figure}

The above sample was the last, and thus least polluted partition, however, the other partitions contained far greater quantities of microplastics and organic material. Below is a zoomed in image of the most concentrated partition. 

\begin{figure}[h!] \centering \includegraphics[width=8cm]{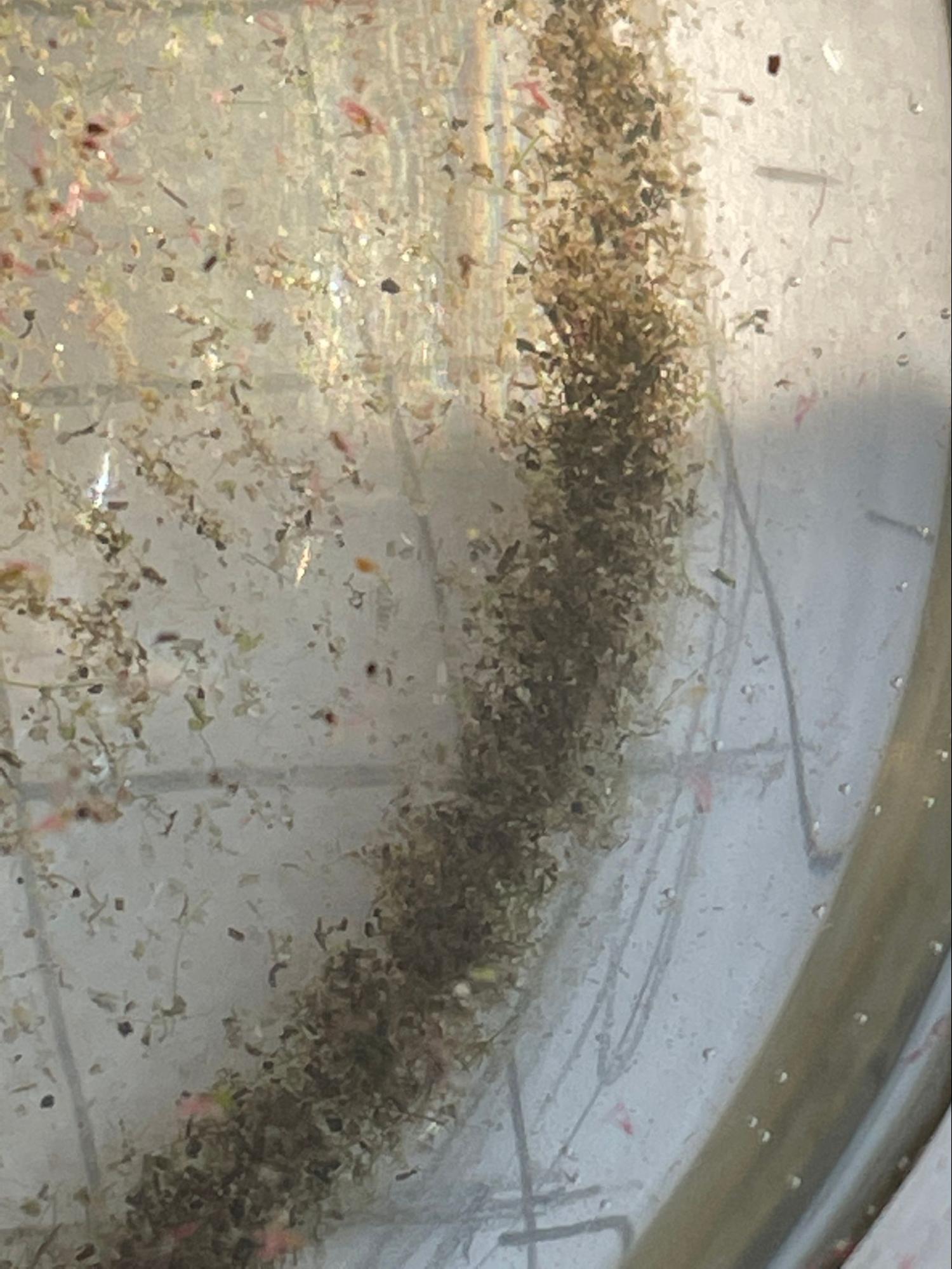}
 \caption{Dense microplastic sample bowl}
\end{figure}

This image is from just after partitioning the plastics, however the plastics in this sample were distributed away from the edge in order to make the counting process less prone to error. In this image, there are many clear, obviously plastic particles, however the image also makes evident how arduous the counting process was for the more concentrated samples. The container shown just above contained the majority of the plastics due to it being the first wash out of the cod-end of the trawl. The spreadsheet of that container is shown below.

\begin{figure}[h!] \centering \includegraphics[width=8cm]{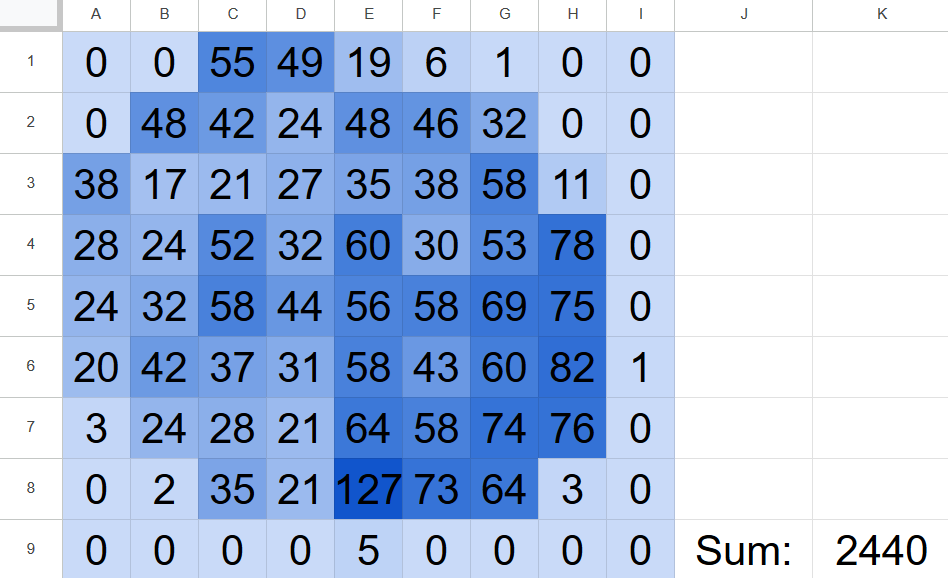}
 \caption{Sparse microplastic sample spreadsheet}
\end{figure}

After counting the plastics, samples were partitioned into metal and glass containers and then frozen so as to preserve them and prevent any organic matter left in the samples from growing. 

\section{Improvements After Testing}
The first test of the trawl in the Milwaukee River was an incredibly enlightening one, and many improvements were made in recognition of the several shortcomings of the trawl. The test was conducted by carrying the trawl to a launch point on the Milwaukee River at Estabrook Park along with two kayaks, one of which had a long rope tied to it and the trawl in order to facilitate retrieval in the worst case scenario of the trawl sinking. The trawl was then carried out on one of the kayaks (a quite difficult task) due to the shallow slope into the river, and turned on once in the water. 

One of the most significant issues faced during the trawl’s testing was that of stability. Although the waters of the Milwaukee River at the location of testing are relatively calm, the trawl still failed to stay completely upright for much of its operation. This was surprising at first, given the very steady state of the craft in a previous test inside of Nicolet Union High School’s swimming pool, however the currents in the river added an unexpected factor which greatly reduced the craft’s stability. Specifically, the currents below the surface and at the surface did not always travel in exactly the same direction or at the same speed. This caused the trawl to begin flipping slowly off-kilter over the course of a few minutes in the river, forcing it to be stabilized every few minutes by a kayak alongside it. In addition to causing issues with propulsion, this instability also exacerbated the already existing issue of the trawl’s control box being poorly mounted onto the top bar of the trawl. This poor mounting arose from the inherent difficulties of mounting a box on a pipe, as it was very difficult to keep the box’s bottom perpendicular to the trawl mouth, and outside forces could relatively easily tip the box in place. This tipping contributed heavily to the second largest issue while testing: that of flooding. 

Due to the lid of the trawl not being water-tight, as soon as the trawl tipped more than \(\approx\)30\(\degree\) backwards, the box began to tip an extra \(\approx\)45\(\degree\), causing it to begin flooding rapidly. This was a massive issue, as the box not only lacked drainage systems but was also filled with sensitive electronics which were not waterproof. Thus in the middle of testing, the trawl had to be lifted out of the water, drained, have its electronics reset, and placed back in while being held by a person in a kayak. This was of course not ideal, especially as the box’s lid was held on by zip-ties which needed to be cut and reset each of the three times which this occurred. In addition to this issue, the motors on the trawl were clogged with algae which needed to be removed several times during the testing process. Finally, the battery voltage lowered to such an extent that it was no longer possible to run the motors at the intended speed after roughly 1 hour and 15 minutes of operation (although this was not necessarily a bad thing). 

Overall, these observations all led to four primary design changes which needed to be made:

\begin{enumerate}
    \item The control box on top of the trawl needed to be completely redesigned, including but not limited to:

\begin{enumerate}
        \item A new, more water-resistant lid which could be held on without the use of zip-ties.
        \item A drainage system to allow water to exit if/when it entered. 

    \begin{enumerate}
            \item Elevated platforms for sensitive electronics so that they could not be damaged by water. 
Water-resistant coating for the electronics to ensure that they weren’t disrupted by mild water contact.

    \end{enumerate}

        \item A new attachment system which included attachment points to the lower bar of the trawl in order to increase stability.
\end{enumerate}

    \item The battery needed to be significantly improved for greater battery life. 

\begin{enumerate}
        \item The original battery with 1300 mAh drained in roughly 1 hour, so the battery capacity was increased by over 11 times to a 15000 mAh battery. 
        \item Solar panels needed to be mounted for greater battery life.
Charging could be done in the daytime (conveniently the period when diel vertical migration causes the greatest concentration of microorganisms to be present at the surface) and operation at night.

\end{enumerate}

    \item The motors needed to have a filter mounted in front of them in order to prevent the entry of algae or other aquatic plants into the trawl.
    \item Weight needed to be added to the base, and buoyancy added to the top of the trawl in order to increase stability when in choppier waters. 
\end{enumerate}
 
In addition to all of these necessary design improvements, one significant improvement of adding a GPS tracking system to the trawl was completed in order to allow for retrieval of stuck trawls and monitoring of operating trawls. These improvements come together to make a truly final product which is ready to be deployed and monitored in a marine environment.

\section{Discussion and Results}

The final product created by this project is a modified Manta Trawl with the theoretical ability to clean bodies of water on a scale evidenced by the models above, thus able to begin the work of cleaning up the Great Lakes of microplastic debris. This trawl is controlled by the device shown below, with central processing power provided by an Arduino Mega 2560 along with a waterproof ultrasonic sensor, propulsion provided by a pair of mounted brushless motors, and electrical power provided by a combination of a 12-volt and 9-volt battery (12v for the motors, 9v for the Arduino and sensor). 

\begin{figure}[h!] \centering \includegraphics[width=8cm]{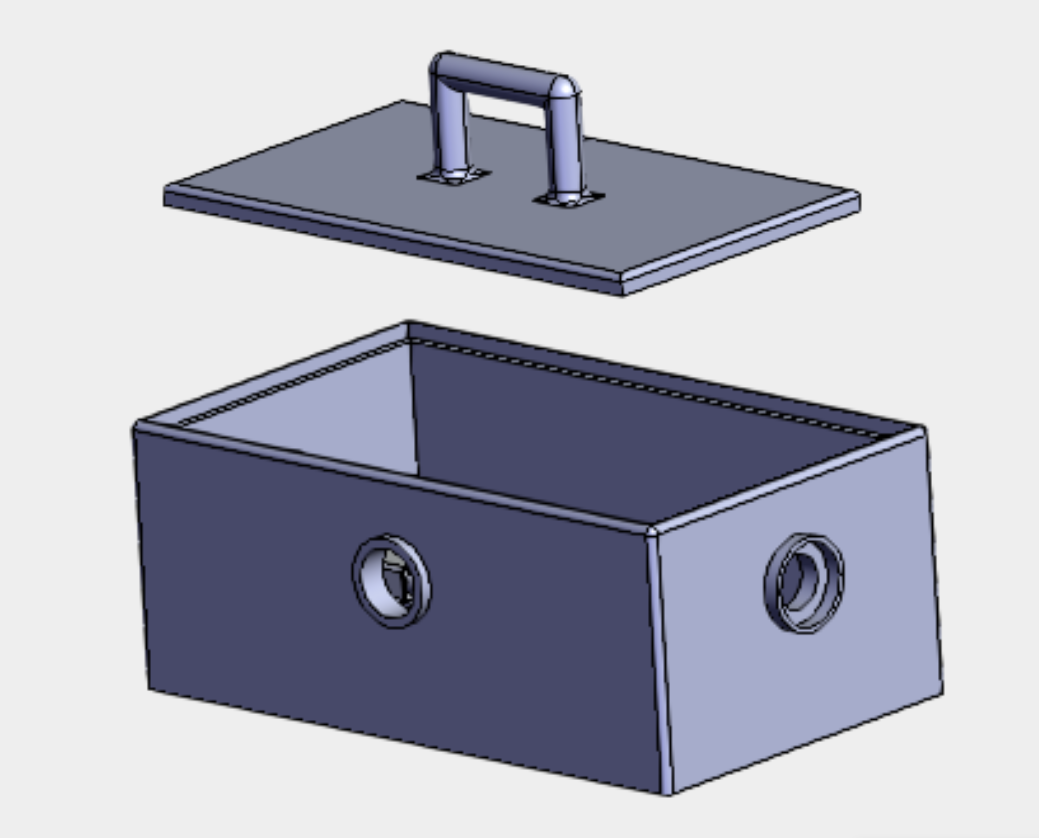}
 \caption{Initial CAD model}
\end{figure}

The image above shows a 3D model of the first iteration of the final product’s control box. This design worked very well in very calm waters, however it, along with the rest of the craft, had several flaws which made it unsuitable for more unstable currents. The primary 4 flaws in the overall design as aforementioned were that:
\begin{enumerate}
    \item The control box was very prone to flooding if tipped.
This, coupled with its inability to drain water once it had entered the box made the first iteration design flood multiple times during testing.

    \item The control box was also very prone to tipping if the trawl tipped more than a few degrees below it.
    \item The trawl was very prone to tipping due to a poor combination of ballast and buoyancy.
    \item The motors became caught with algae relatively easily, alternately reducing their efficiency and preventing them from spinning altogether. 
\end{enumerate}

 In addition to these flaws with the box itself, there were also problems with the battery life of the device, with the primary battery falling from 12.6v to 5.6v over the course of the hour of testing. After reaching roughly 7-8 volts, the motors began to noticeably slow down, an issue which was further exacerbated by the algae getting stuck in the motors. Upon first observation it appeared that the issues with algae only began roughly halfway through the testing, however it may be just as likely that when at full power, the motors were able to simply expel the algae instead of getting stuck on it –this would have led to the same observation of algae problems beginning just past halfway through the testing. Overall, the testing exposed several flaws with the trawl’s design, however it was also quite successful in the trawl’s overall goal of collecting microplastics at a rate equal to or greater than previous studies’ results.
 To fix the shortcomings of the first design, the entire 3D-printed section of the device was redesigned to make several additions and improvements on the previous design. Several changes were also made to the buoyancy of the trawl to attempt fixing some of the issues found in that regard. The major improvements are as follows: 
\begin{enumerate}
    \item Dedicated spaces for solar panels to sit and connect to the internal electronics of the device.
    \item Complete change in the system attaching the control vessel to the trawl from a clamp held by zip-ties to a rigid attachment on both top and bottom of the trawl.
    \item Addition of more ballast and buoyancy to the trawl to make it self-stabilizing even in slightly turbulent conditions.
    \item Moving of ultrasonic sensor location to below the water in order to more effectively detect underwater obstacles.
    \item Addition of drainage to the control box to deal with issues of flooding. 
\end{enumerate}
\begin{figure}[h!] \centering \includegraphics[width=8cm]{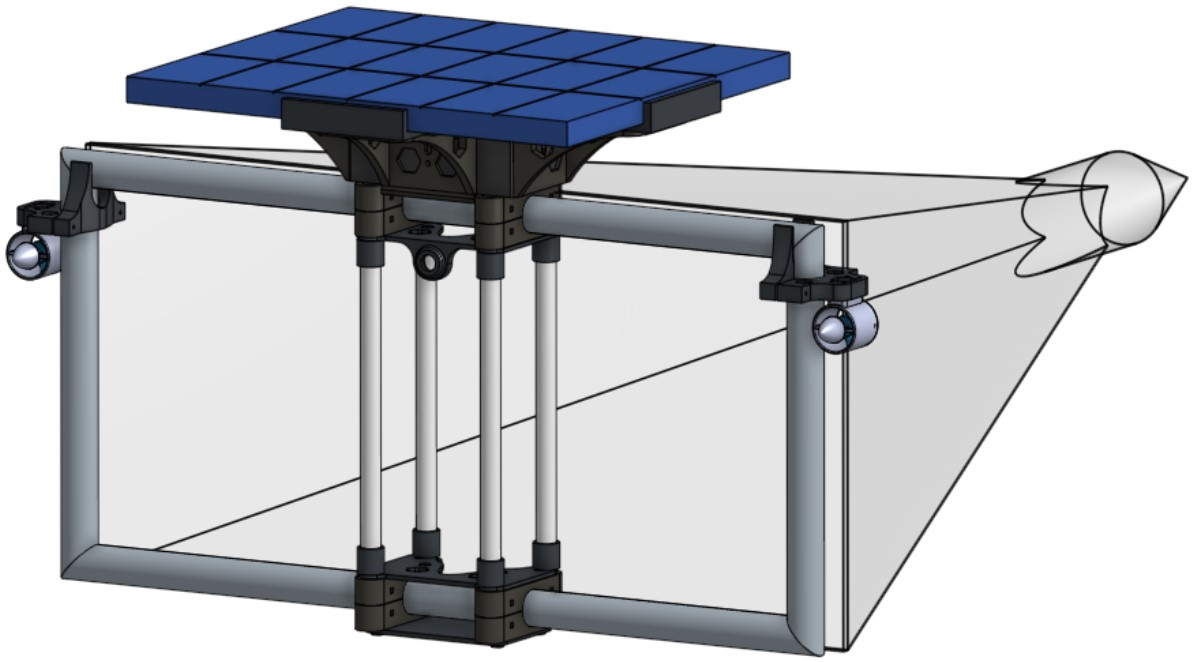}
  \caption{Full 3D model}
\end{figure}

The solar panels which were added are able to recharge the battery within 4 peak sun hours. This works quite well as the Great Lakes see roughly this amount of sunlight on average across the year, with Chicago and Milwaukee seeing 4.0 each, Toronto and Rochester seeing 3.9, and Toledo seeing 4.1. There is a lack of data readily available regarding this on the lakes themselves, however, it can be inferred that the cloud cover is similar to surrounding areas –suggesting that the peak sun hours would be very similar. 

Deploying these trawls, as explained previously, would be a significant endeavor in and of itself, however the process could be aided by following microplastic distribution patterns and lake current maps, by using models to predict where trawls would be most effective, and by studying where microplastics have the greatest effect per unit of concentration in order to affect the greatest change possible and in order to collect plastics most efficiently. As stated in the cost analysis, the trawl created costs roughly \$1,500-\$2,000, and thus deploying on a large scale would cost a relatively significant amount. If deployed at a rate of one per week (and if one is willing to wait for roughly 15 years), the total cost of materials would amount to roughly \$810,000, not accounting for bulk purchases likely cheapening that cost. If deployed at a rate of one per day, again as shown in previous models, the total cost of materials would amount to roughly \$2.4 million, a far more significant sum (both figures for Lake Erie exclusively). These two numbers, in addition to the cost of using boats to deploy these trawls into the lake every day or every week, mean that for this solution to be implemented, significant capital would have to be invested, however this cost still seems rather reasonable when discussing a matter as significant as this. Overall, the presented device is in a state ready for deployment on the Great Lakes given minimal extra testing. 

\section{Next Steps}
The next, and ideally final, step of testing is that of a long-term test on the Great Lakes as the ultimate showcase of each of the systems on the trawl operating in concert with each other so as to prove the long-term viability of the presented device. Inclement weather has thus far prevented this type of testing, however, the device will be tested in the near future once temperatures rise enough that the device can be safely deployed into the lake without risk of temperature related ailment.

\section{Conclusion}
In conclusion, the modified Manta Trawl created in this project, although untested at scale, could in theory significantly contribute to environmental conservation through the removal of microplastics from the Great Lakes (and other bodies of water in the future), and although the system currently developed is but a prototypical one, and a final solution could implement a larger trawl to more effectively collect microplastics, it lends credence to the idea of autonomously cleaning large bodies of water on a large scale, one which has been dismissed by many groups as impossible. Indeed, it is undoubtable that a solution like this is nearly useless if policies are not put into place to limit the introduction of microplastics into aquatic environments. Without governmental changes, there is simply no way for any solution focussed on cleaning up lakes, rivers, or oceans to be feasible in the very long term. Even today, despite widespread environmental activism, the number of plastics deposited into the worlds’ bodies of water grows rapidly, and every year concerning reports on the state of the world’s waterways come and go without meaningful change in how governments approach the issue of plastic waste. Thus, this solution can only be a temporary one, and although it is a particularly effective one, it cannot last forever. This issue cannot be fixed in one fell swoop, and one can only hope that in the near future, action is taken to definitively curb the plastic pollution which pollutes our waterways each day. 

\section{Declarations}
The author has no financial, competing interest, or personal interests to declare. 


\phantomsection
\bibliographystyle{unsrt}
\bibliography{main.bib}
\nocite{*}


\end{document}